\documentclass{article} 
\usepackage{iclr2026_conference,times}
\usepackage{graphicx}
\usepackage{subcaption}
\usepackage{booktabs}
\usepackage{wrapfig}
\usepackage{multirow}
\usepackage[colorlinks=true, linkcolor=blue, citecolor=blue, urlcolor=blue]{hyperref}
\usepackage{amsthm} 
\newtheorem{theorem}{Theorem}
\usepackage[table]{xcolor}

\usepackage{amsmath,amsfonts,bm}

\def\eqref#1{equation~\ref{#1}}

\def\1{\bm{1}}

\DeclareMathAlphabet{\mathsfit}{\encodingdefault}{\sfdefault}{m}{sl}
\SetMathAlphabet{\mathsfit}{bold}{\encodingdefault}{\sfdefault}{bx}{n}

\usepackage{hyperref}
\usepackage{url}

\usepackage{ifthen}
\usepackage{xcolor}

\title{Understanding and Enhancing the Planning Capability of Language Models via Multi-Token Prediction}

\author{Qimin Zhong$^{1}$, Hao Liao$^{1}$, Siwei Wang$^{2}$, Mingyang Zhou$^{1}$ \\\textbf{Xiaoqun Wu$^{1}$, Rui Mao$^{1}$, Wei Chen$^{2}$} \\
$^{1}$College of Computer Science and Software Engineering, Shenzhen University, China \\
$^{2}$Microsoft Research Asia}

\iclrfinalcopy

\begin{document}

\maketitle

\begin{abstract}
Large Language Models (LLMs) have achieved impressive performance across diverse tasks but continue to struggle with learning transitive relations, a cornerstone for complex planning~\citep{Balesni2024twohopcurse,wang2024alpine}. To address this issue, we investigate the Multi-Token Prediction (MTP) paradigm and its impact to transitive relation learning. We theoretically analyze the MTP paradigm using a Transformer architecture composed of a shared output head and a transfer layer. 
Our analysis reveals that the transfer layer gradually learns the multi-step adjacency information, which in turn enables the backbone model to	capture unobserved transitive reachability relations beyond those directly present in the training data, albeit with some inevitable noise in adjacency estimation. Building on this foundation, we propose two strategies to enhance the transfer layer and overall learning quality: \emph{Next-Token Injection (NTI)} and a \emph{Transformer-based transfer layer}. Our experiments on both synthetic graphs and the Blocksworld planning benchmark validate our theoretical findings and demonstrate that the improvements significantly enhance the model’s path-planning capability. These findings deepen our understanding of how Transformers with MTP learn in complex planning tasks, and provide practical strategies to overcome the transitivity bottleneck, paving the way toward structurally aware and general-purpose planning models.
\end{abstract}
\section{Introduction}
Transformer models have achieved remarkable success across natural language processing~\citep{vaswani2017attention,devlin2019bert,brown2020language}, computer vision~\citep{dosovitskiy2020image,carion2020end,liu2021swin}, reinforcement learning~\citep{parisotto2020stabilizing,chen2021decision,janner2021offline}, program synthesis~\citep{chen2021evaluating,nijkamp2022codegen}, and complex planning~\citep{chen2021decision,lehnert2024beyond}. However, a fundamental question remains: do these models truly possess planning capabilities, or do they merely rely on reconstructing patterns from training data? This question is particularly critical in complex planning tasks, which often require compositional planning to generate coherent sequences of actions toward a goal. In such tasks, it is natural and effective to abstract the problem as path finding on a graph, where nodes represent states and edges represent executable actions. path finding not only lies at the core of many classical planning problems but also closely relates to sequential decision-making in real-world complex tasks, such as robotic motion planning, automated scheduling, and step-wise reasoning in mathematical proofs. Under this abstraction, standard autoregressive Transformers typically perform reliable planning on paths observed during training~\citep{wang2024alpine}.

However, the performance of these models degrades substantially when the task requires transitive planning, which demands combining information from multiple path segments to infer new reachability relations: as demonstrated by~\cite{wang2024alpine}, standard autoregressive Transformers would fail to infer that node $A$ could reach node $C$ when the training data contain both paths from $A$ to $B$ and from $B$ to $C$ but no paths from $A$ to $C$. This limitation not only prevents the model from generalizing to unseen paths in complex planning tasks but also highlights a fundamental bottleneck of current Transformers in structured planning~\citep{zhang2024transformer}. Therefore, understanding and improving the model's compositional learning ability is crucial for enhancing Transformers in structured planning and sequential prediction tasks.

To address this issue, we explore the Multi-Token Prediction (MTP) paradigm, in which the model predicts multiple future nodes in a single training step, providing richer supervision signals and showing potential for modeling long-range dependencies and structural relationships. MTP has been adopted in a number of leading AI companies and their models such as Meta and DeepSeek~\citep{gloeckle2024better,liu2024deepseek}, but their underlying mechanism, especially for planning, remain largely unexplored. In this work, we build on the analytical framework of ALPINE~\citep{wang2024alpine} to systematically investigate the effect of MTP on path planning, and propose architectural enhancements to strengthen the ability of Transformers to learn transitive reachability relations. Our study provides both theoretical insights and practical guidance for developing future Transformers with stronger reasoning and planning capabilities.

To summarize, our contributions include:
First, through a theoretical analysis on a simplified Transformer, we show how the multi-token loss simultaneously shapes the transfer matrices and backbone network weights, revealing the coupled learning dynamics among the transfer layer and the adjacency and reachability within the backbone model (Section~\ref{methods}). Second, based on these insights, we propose two enhancements to the architecture: (1) \emph{Next-Token Injection (NTI)}, to explicitly inject intermediate nodes as multi-hop supervision; and (2) a \emph{Transformer-based transfer layer} to maintain structural consistency across prediction steps  (Section~\ref{subsec:transfer_strategy}). Third, we conduct extensive experiments on synthetic graphs as well as the Blocksworld planning benchmark and show that these methods significantly improve prediction accuracy, and the learned transfer matrices progressively approximate the ground-truth adjacency matrices (Section~\ref{sec:experiments}). These findings indicate that MTP together with our enhancements provides better support for transitive relations,  advancing models toward stronger structural planning capabilities.

\paragraph{\bf Additional Related Work}
Our work relates to recent studies on the structural planning ability of large language models (LLMs). Prior work has examined LLMs in path prediction and task planning over symbolic or graph-structured inputs. Other approaches use prompt engineering or structural injection to guide graph reasoning. Multi-Token Prediction (MTP) has been explored to improve training efficiency. Unlike these, we focus on the interpretability of structural learning under MTP. See Appendix~\ref{appendix:related-work} for more details.

\section{Setting and Preliminaries}
\label{preliminaries}
We use $\bm{a}$ to denote a column vector, $\bm{A}$ denotes a matrix; the $i^{th}$ component of $\bm{a}$ is written as $\bm{a}_{(i)}$; the $(i,j)$ entry of $\bm{A}$ is written $\bm{A}_{(i,j)}$; the $i$-th row (column) of $\bm{A}$ is denoted $\bm{A}_{(i,:)}$ ($\bm{A}_{(:,i)}$).

\subsection{Path Planning over Simple Directed Graphs with Language Model}

To evaluate the planning capability of an autoregressive language model, we construct path-planning tasks on directed graphs. Let \(\mathcal{G} = (\mathcal{V}, \mathcal{E})\) be a directed acyclic graph with node set \(\mathcal{V}\) and edge set \(\mathcal{E}\). For any \(u, v \in \mathcal{V}\), the presence of \((u,v) \in \mathcal{E}\) indicates a directed edge from \(u\) to \(v\).

During training, each reachable source–target pair \((s,t)\) (i.e., \(t\) can be reached from \(s\) via one or more edges) is encoded as a token sequence \(\texttt{"s t s a b c t \textbackslash n"}\), where 
 \(s\) and \(t\) denote the source and target, \(a,b,c\) represent intermediate nodes, and \(\texttt{\textbackslash n}\) marks sequence termination. The model learns in an autoregressive fashion by predicting every next token in turn.
At test time, only the prefix \(\texttt{"s t"}\) is provided; the model must autoregressively complete a valid path from \(s\) to \(t\), respecting the graph’s adjacency and reachability constraints. This procedure measures the model’s ability to capture both one-step adjacency and long-range reachability information.

We denote the ground‐truth adjacency matrix and reachability matrix of the graph by \(\bm{A}^{\mathrm{true}}\) and \(\bm{R}^{\mathrm{true}}\), respectively:
\[
\bm{A}^{\mathrm{true}}_{(i,k)} = 
\begin{cases}
1, & \text{if } (i,k)\in\mathcal{E},\\
0, & \text{otherwise}.
\end{cases}
\quad
\bm{R}^{\mathrm{true}}_{(t,k)} = 
\begin{cases}
1, & \text{if } \text{there exists a path }k\to t\text{ in }\mathcal{G},\\
0, & \text{otherwise}.
\end{cases}
\]

\subsection{Hierarchical Evaluation of Generalization Ability}

Given a training set \(\mathcal{D}\) where 
	$\bm{u}=(u_1,\dots,u_n)\in \mathcal{D}$ is a valid path in the graph, 
	we define the observed adjacency and reachability matrices in the observation graph \(\mathcal{G}_{\mathrm{obs}}\):
\[
\begin{aligned}
\bm{A}^{\mathrm{obs}}_{(i,k)} &= 
\begin{cases}
1, & \text{if } \exists\,\bm{u}\in\mathcal{D},\,n\in[3,N-1]\ \text{s.t. } u_n = i,\ u_{n+1} = k, \\
0, & \text{otherwise}
\end{cases} \\
\bm{R}^{\mathrm{obs}}_{(t,k)} &= 
\begin{cases}
1, & \text{if } \exists\,\bm{u}\in\mathcal{D},\,n\in[4,N]\ \text{s.t. } u_2 = t,\ u_{n} = k, \\
0, & \text{otherwise}.
\end{cases}
\end{aligned}
\]
Here, $\bm{R}^{\mathrm{obs}}$ is a subset of $\bm{R}^{\mathrm{true}}$, contains only the reachability relations directly observed in \(\mathcal{D}\).

We then partition test pairs \((s,t)\) into four \emph{degrees} based on their observed reachability in \(\mathcal{G}_{\mathrm{obs}}\): a) degree-0 if \(\bm{R}^{\mathrm{obs}}_{(t,s)}=1\); b) degree-1 if \(\bm{R}^{\mathrm{obs}}_{(t,s)}=0\) but there exists \(u\) such that \(\bm{A}^{\mathrm{obs}}_{(s,u)}=1\) and \(\bm{R}^{\mathrm{obs}}_{(t,u)}=1\); c) degree-2 if it is neither degree-0 nor degree-1 but there exists \(u\) such that \(\bm{A}^{\mathrm{obs}}_{(s,u)}=1\) and \((u,t)\) is degree-1; d) degree-3 otherwise.

Following the standard architecture of Generative Pretrained Transformer (GPT)~\citep{radford2018improving},
each Transformer layer comprises multi‐head attention (MHA), residual connections, layer normalization (LN), and a feed-forward network (FFN), as
\begin{equation}
\mathrm{Transformer}(\bm{X})
=\mathrm{FFN}\bigl(\mathrm{LN}_2(\mathrm{MHA}(\mathrm{LN}_1(\bm{X}))+\bm{X})\bigr)
+\mathrm{MHA}(\mathrm{LN}_1(\bm{X}))+\bm{X}.
\label{eq:transformer_full}
\end{equation}
A training sequence 
$\bm{u}=(u_1,\dots,u_n)$
is first mapped to a sequence of corresponding embedding vectors 
$\bm{X}_{1:n}=(\bm{x}_1,\dots,\bm{x}_n)$
using an embedding matrix $\bm{W}_t$.
This sequence is then passed through the Transformer layers, which produce a sequence of contextualized hidden states
\begin{equation}
\bm{H}_{1:n}=(\bm{h}_1,\dots,\bm{h}_n)=\mathrm{Transformer}(\bm{X}_{1:n}).
\end{equation}
For next-token prediction, the model uses the final hidden state $\bm{h}_n$, which corresponds to the last token in the input sequence.
The predictive distribution is then given by:
\[
\bm{p}(u_{n+1} \mid u_{1:n}) = \mathrm{softmax}(\bm{W}_o \bm{h}_n).
\]
The standard next-token training objective is the cross-entropy loss,
\begin{equation}
\mathcal{L}
=-\sum_{n=1}^{N-1}\log \bm{p}(u_{n+1}\mid u_{1:n}).
\end{equation}
\cite{wang2024alpine} point out that 
	GPT trained solely with next-token loss achieves over 90\% accuracy on degree-0 and degree-1 tasks, but drops to about 60\% on degree-2 tasks. The model can only learn the observed reachability matrix \(\bm{R}^{\mathrm{obs}}\), and fails to learn the complete true reachability matrix \(\bm{R}^{\mathrm{true}}\), highlighting its inability to generalize to transitive paths unseen during training.

\section{Mechanistic Understanding of Multi-Token Prediction}
\label{methods}

\begin{figure}[htbp]
    \centering
    \includegraphics[width=0.85\textwidth]{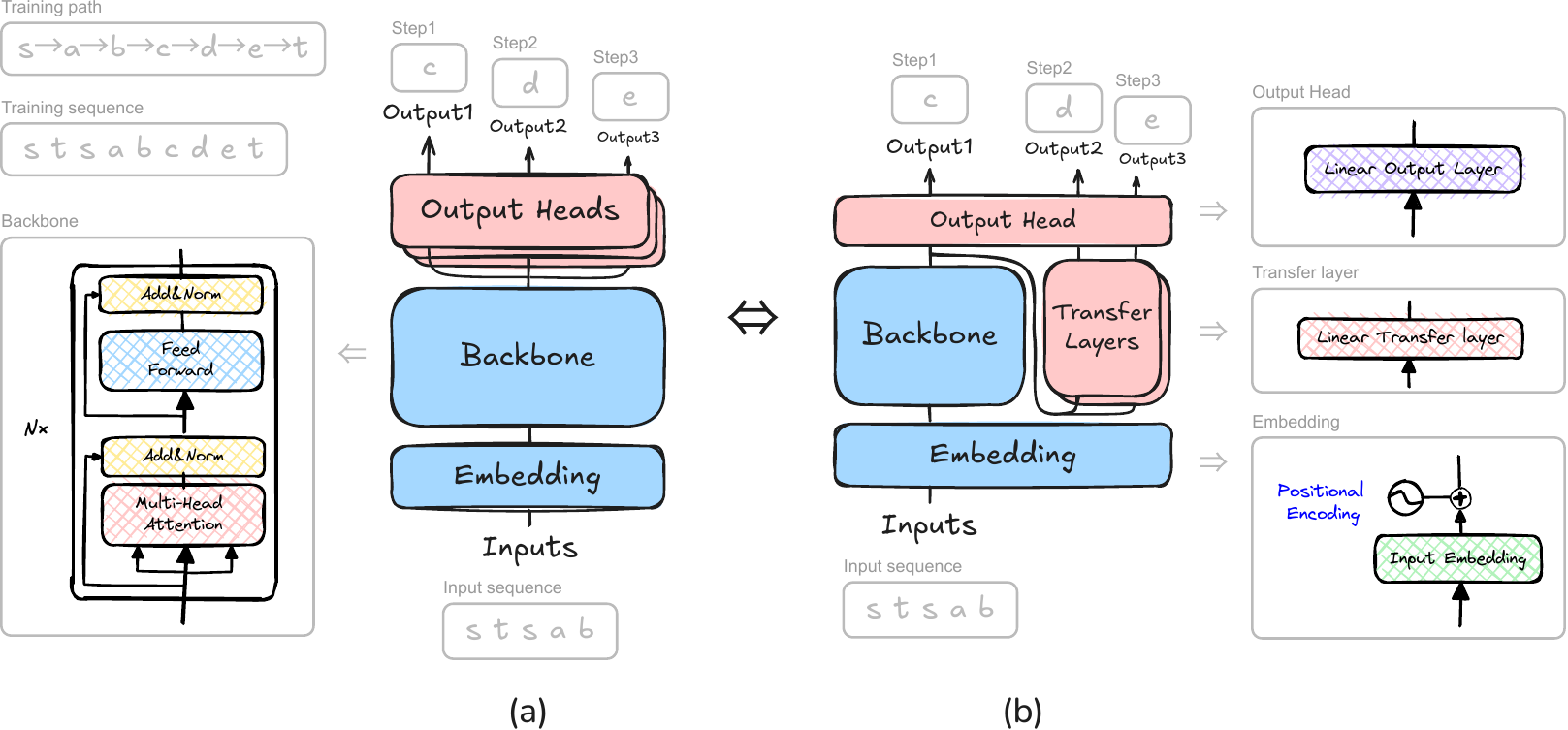}
    \caption{
    \textbf{Multi-Token Prediction (MTP) architectures for 3-step prediction.}  
    (a) Meta's MTP architecture with independent output heads for each step.  
    (b) Ours with a shared output head:  
    \quad i) next-step predictions are generated directly from the backbone,  
    \quad ii) other steps require transformation through a separate transfer layer.
    }
    \label{fig:unified_structure}
\end{figure}

In this paper, we investigate the mechanism of MTP in which multi-step tokens are used during learning to enhance the learning effectiveness, while 
	inference is still done by next-token prediction.
Some prior work also uses MTP at inference to accelerate generation, but it is not our focus (See  Appendix~\ref{appendix:related-work} for more discussions).
We use a shared output head architecture for MTP (Figure~\ref{fig:unified_structure}(b)). Separate learnable {\em transfer layers} map the backbone output to different target positions in parallel before decoding, enabling the model to share parameters across steps. 
This design enhances the interpretability of structural learning. 
The shared output head is denoted as \(\bm{W}_o\), and the transfer layer as \(\bm{W}^T\).
In contrast, Meta’s MTP architecture assigns an independent output head to each prediction step~\citep{gloeckle2024better} (Figure~\ref{fig:unified_structure}(a)). While flexible, this approach lacks parameter sharing, making it harder to model unified patterns across steps. 
The two architectures are mathematically equivalent and thus we choose the first one for its better interpretability.

\subsection{Training Dynamics of 2-Token Prediction}
\label{4.2}

To enable theoretical analysis on how 2-token supervision shapes structural learning, we simplify the Transformer model with several assumptions, similar to the ones in~\citep{wang2024alpine}. 
We consider a single-layer, single-head Transformer where positional embeddings and layer normalizations are omitted; the feed-forward network is a single linear map $\mathrm{FFN}(\bm X)=\bm X \bm W^M$; and the token embedding matrix $\bm{W}_t$ and output embedding matrix $\bm{W}_o$ are identity matrices. The attention mechanism is also simplified, using a standard value projection matrix $\bm{W}^V$ but with a manually set attention matrix $\bm{\alpha}$ (replacing the standard $\mathrm{softmax}\left(\frac{\bm{Q K}^\top}{\sqrt{d_k}}\right)$) that restricts attention to the target token (i.e., each row is a one-hot vector with a 1 in the second column). Finally, a transfer matrix $\bm W^T \in \mathbb{R}^{M \times M}$ maps next-step logits to subsequent-step logits. Complete derivations under these assumptions are provided in Appendix~\ref{appendix:toy-analysis}.

Under this setup, the hidden state at position $n$, $\left( \bm{H}_1 \right)_{(n,:)}$, is the sum of feed-forward and attention outputs derived from the one-hot input matrix $\bm{U}$. Projecting this state through the output matrix $\bm{W}_o$ yields the logits:
\[
\left( \bm{H}_1 \right)_{(n,:)} \bm{W}_o 
= \left( \bm{U} \bm{W}_t \bm{W}^M + \alpha \bm{U} \bm{W}_t \bm{W}^V \right)_{(n,:)} \bm{W}_o
= \left( \bm{U} \bm{W}^M + \alpha \bm{U} \bm{W}^V \right)_{(n,:)}
= \bm{W}^M_{(u_n,:)} + \bm{W}^V_{(u_2,:)},
\]
where $u_n$ is the current token and $u_2$ is the attended target token. Therefore, the logits for the next and the subsequent step are given by
\[
\mathrm{logit}_{n+1}(k)=\bm W^M_{(u_n,k)}+\bm W^V_{(u_2,k)},\quad
\mathrm{logit}_{n+2}(k)=(\bm W^M\bm W^T)_{(u_n,k)}+(\bm W^V\bm W^T)_{(u_2,k)}.
\]

For 2-Token Prediction, the objective is $\ell(\mathcal{D})=\ell^{(1)}(\mathcal{D})+\ell^{(2)}(\mathcal{D})$, and we focus on the second loss $\ell^{(2)}(\mathcal{D})$, which is formulated as:
\begin{equation}
\ell^{(2)}(\mathcal{D}) = 
- \sum_{u \in \mathcal{D}} \sum_{n=1}^{N-2} \sum_k \bm{U}_{(n+2,k)} \log
\frac{
\exp\left( (\bm{W}^M \bm{W}^T)_{(u_n,k)} + (\bm{W}^V \bm{W}^T)_{(u_2,k)} \right)
}{
\sum_{\ell} \exp\left( (\bm{W}^M \bm{W}^T)_{(u_n,\ell)} + (\bm{W}^V \bm{W}^T)_{(u_2,\ell)} \right)
}.
\end{equation}

Let $\hat{P}_{i,j}(k')$ be the softmax probability of predicting node $k'$ two steps ahead given current node $i$ and target $j$. Define $N_{i,j,k'}$ as the number of such occurrences in $\mathcal{D}$ and $N_{i,j}=\sum_{k'}N_{i,j,k'}$. This leads to the following theorem.

\begin{theorem}\label{thm:1}
For any pair \((i, j)\) in dataset \(\mathcal{D}\) with \(N_{i,j} > 0\), let \(P^{\mathrm{data}}_{i,j}(k') = \frac{N_{i,j,k'}}{N_{i,j}}\) be the empirical probability of the second-next node \(k'\). The contribution of this pair to the gradient \(\frac{\partial \ell^{(2)}(\mathcal{D})}{\partial \bm{W}^T_{(d,k')}}\) is determined by the prediction error, for any \(d\) where \((\bm{W}^M_{(i,d)} + \bm{W}^V_{(j,d)}) > 0\):
(i) If \(\widehat{P}_{i,j}(k') < P^{\mathrm{data}}_{i,j}(k')\), the contribution is \textbf{negative}, promoting an \textbf{increase} in the weight \(\bm{W}^T_{(d,k')}\).
(ii) Conversely, if \(\widehat{P}_{i,j}(k') > P^{\mathrm{data}}_{i,j}(k')\), the contribution is \textbf{positive}, promoting a \textbf{decrease} in the weight.
The total gradient is the sum of these contributions over all pairs \((i, j)\) in \(\mathcal{D}\).
\end{theorem}

The derivations and proofs of the theorems are provided in Appendix~\ref{appendix:gradient-analysis}. 

\paragraph{Transfer Matrix Learned as an Adjacency Matrix.}  
Theorem~\ref{thm:1} shows that the transfer matrix \(\bm{W}^{T}\) is updated by 2nd-step prediction errors. When the model underestimates the probability of reaching node \(k'\) from node \(i\) in two steps, the gradient increases the weight \(\bm{W}^{T}_{(d,k')}\) from all the positive-correlated intermediate node \(d\) (e.g., all the feasible $d$'s for this $i,j$ pair) to \(k'\); otherwise, it decreases. Thus, if the backbone model correctly predicts the next-step node \(d\), then by increasing the weight \(\bm{W}^{T}_{(d,k')}\), it will enable \(\bm{W}^{T}\) to correctly learn the adjacency between \(d\) and \(k'\).

We next analyze how the 2nd-step prediction affects the backbone parameters $\bm{W}^M$ and $\bm{W}^V$ through gradients propagated from the transfer matrix.

\begin{theorem}\label{thm:2}  
For any pair \((i, j)\) in dataset \(\mathcal{D}\) with \(N_{i,j} > 0\), the contribution of each \((\text{current node } i, \text{second-step node } k')\) pair to the gradient \(\frac{\partial \ell^{(2)}(\mathcal{D})}{\partial \bm{W}^V_{(j,k)}}\) is determined by the prediction error, for any \(k\) where \(\bm{W}^T_{(k,k')} > 0\):  
(i) If \(\widehat{P}_{i,j}(k') < P^{\mathrm{data}}_{i,j}(k')\), the contribution is \textbf{negative}, promoting an \textbf{increase} in the weight \(\bm{W}^V_{(j,k)}\);  
(ii) Conversely, if \(\widehat{P}_{i,j}(k') > P^{\mathrm{data}}_{i,j}(k')\), the contribution is \textbf{positive}, promoting a \textbf{decrease} in the weight.  
The total gradient is the sum of contributions from all \((i, k')\) pairs.  
Analogous results hold for gradients w.r.t.\ \(\bm{W}^M\).  
\end{theorem}

\paragraph{Learning the Transitive Reachability.}  
The next-token loss $\ell^{(1)}(\mathcal{D})$ encourages the backbone matrix $\bm{W}^V$ to capture the observed reachability from the training data~\citep{wang2024alpine}. 
For a given pair $(i,k')$, when the transfer matrix entry $\bm{W}^T_{(k,k')}$ is large (indicating a confident $k \rightarrow k'$ transition), and the model predicts a lower probability for $k'$ than the ground truth along a path $i \rightarrow k \rightarrow k'$, the 2nd-step prediction loss $\ell^{(2)}(\mathcal{D})$ applies a negative gradient to $\bm{W}^V_{(j,k)}$ (i.e., increasing its weight), thereby strengthening the $k \rightsquigarrow j$ relation.
Therefore, when $\bm{W}^T_{(k,k')}$ captures the true adjacency relationship between $k$ and $k'$, the 2nd-step prediction enables the backbone to learn the {\em transitive reachability} from $k$ to $j$, based on the observed reachability from $k'$ to $j$ is learned by $\bm{W}^V_{(j,k')}$ by the 1st-step token prediction, and the adjacency $(k,k')$ is learned by the transfer layer $\bm{W}^T_{(k,k')}$.
This shows that 2-token prediction could achieve higher-order reachability beyond the observed reachability of the next-token prediction.

\paragraph{Learning the Adjacency.} 
While the next-token loss $\ell^{(1)}(\mathcal{D})$ directly encourages $\bm{W}^M$ to capture the adjacency relationship in the dataset~\citep{wang2024alpine}, the 2nd-step prediction loss $\ell^{(2)}(\mathcal{D})$ operates indirectly. For a given pair $(j,k')$, when $\bm{W}^T_{(k,k')}$ is large and the model underestimates the probability of the second-step node $k'$, the loss applies a negative gradient to $\bm{W}^M_{(i,k)}$, thereby strengthening the $i \rightarrow k$ connection. This suggests that {\em{spurious adjacency}} $(i,k)$ may be introduced into $\bm{W}^M$ when learning transitive reachability by the 2nd-step prediction, which is mechanically difficult to avoid due to the tight coupling between adjacency and reachability learning in the backbone. Our empirical validation later (Section~\ref{sec:experiments}) demonstrates that this risk is limited and the overall benefit of learning transitive reachability outweighs the risk of spurious adjacency. 

\paragraph{Next-Node Prediction.}   During the next-token prediction inference, the model samples the next node $k$ with high $\bm{W}^M_{(u_n, k)} + \bm{W}^V_{(u_2, k)}$,
	which favors nodes that are both neighbors of the current node (high $\bm{W}^M$) and reachable from the target (high $\bm{W}^V$), and correctly reflects the essence of path planning. Moreover, the transitive reachability learned from $\ell^{(2)}$ helps the model generate more accurate paths, leading to an improvement on its performance, especially for high-order test cases. 

\subsection{Learning Mechanism of Multi-Token Prediction}  

The total loss of MTP is defined as the sum of cross-entropy losses at each step: \(\ell(\mathcal{D}) = \sum_{k=1}^{n} \ell^{(k)}(\mathcal{D}),\) where each \(\ell^{(k)}\) corresponds to an independent transfer layer. The transfer layer used for generating the outputs of the $n$-th step token is denoted as \(\mathbf{W}^{T{(n-1)}}\). 

\begin{figure}[t]
    \centering
    \includegraphics[width=0.85\textwidth]{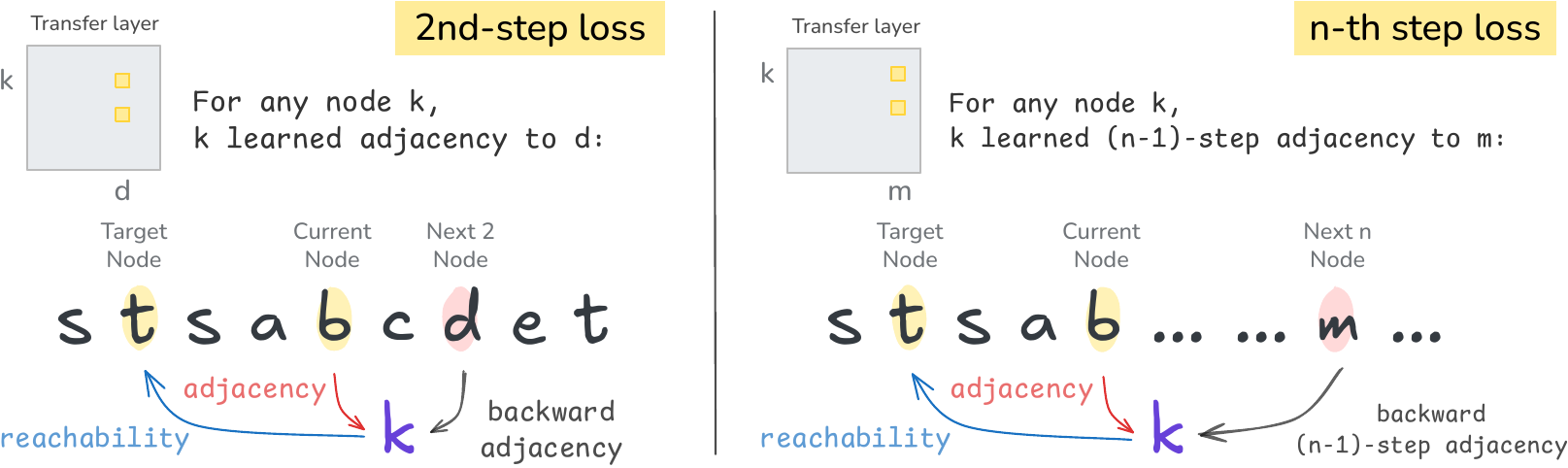}
    \caption{
    \textbf{Illustration of the learning mechanism under Multi-Token Prediction.}
    Left: 2nd-step loss; Right: $n$-th step loss. MTP learns transitive reachability and spurious adjacency.}
    \label{fig:token_loss_mechanism}
\end{figure}

Theorems~\ref{thm:1} and~\ref{thm:2} generalize naturally: 
	the \(n\)-th transfer layer \(\bm{W}^{{T}{(n-1)}}\) takes the next-step logits from the backbone model as input and outputs 
	the \(n\)-th step logits, such that \(\bm{W}^{{T}{(n-1)}}\) approximates the \((n-1)\)-th power of the adjacency matrix. Under the influence of \(\bm{W}^{{T}{(n-1)}}\), 
	the model can capture the transitive reachability composed of the observed reachability from the $n$-th step node $m$ to target $t$ and
		the $(n-1)$-th power adjacency from some node $k$ to $m$ learned under \(\bm{W}^{{T}{(n-1)}}\) (Figure~\ref{fig:token_loss_mechanism}).
	Meanwhile, it may learn some spurious adjacency from the current node $b$ to $k$.

\section{Enhancing Transfer Layers for Multi-Token Prediction}
\label{subsec:transfer_strategy}

\begin{wrapfigure}{r}{0.40\textwidth}
    \centering
    \vspace{-34pt}
    \includegraphics[width=0.38\textwidth]{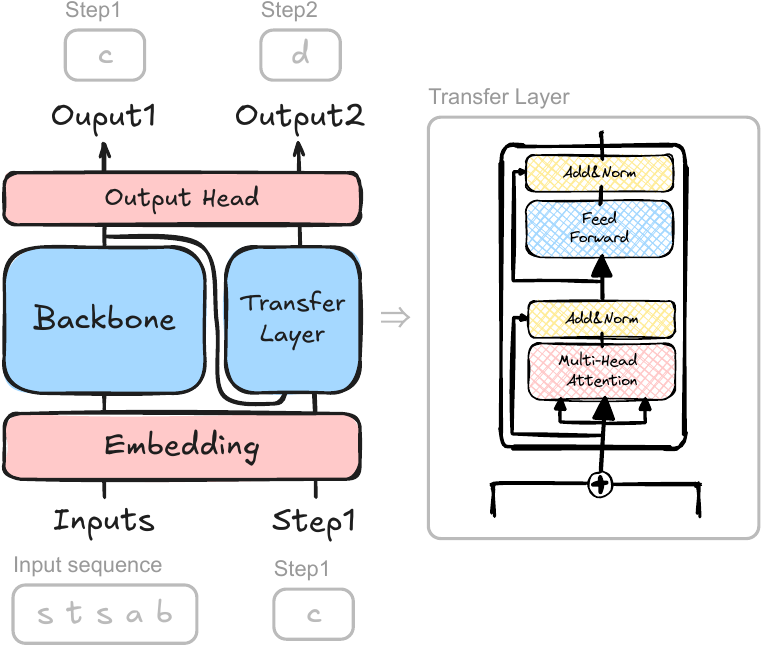}
    \caption{ 
    Enhanced transfer layer architecture with NTI and Transformer-based transfer layer.}
    \label{fig:fig_NTI_transfer_arch}
    \vspace{-10pt}
\end{wrapfigure}

\subsection{Next-Token Injection}

The transfer layer projects the backbone representation at the next-step position to predict future tokens. Its performance is constrained by the backbone output $\bm{h}_n$'s ability to predict the next node, where the deviation between the predicted and true next step introduces noise.

We propose \textbf{Next-Token Injection} (NTI), which augments the transfer input with information from the true next node to provide direct supervision. 
This is achieved by injecting the embedding vector of the true next token $u_{n+1}$ into the backbone's hidden state $\bm{h}_n$, which is then mapped to different positions by separate transfer layers, as follows:
\begin{equation}
\widetilde{\bm{h}}_n = \bm{h}_n + k (\bm{W}_t)_{(:,u_{n+1})}, \ 
\operatorname{logits}_{2} = \bm{W}_o\widetilde{\bm{h}}_n \bm{W}^T , 
\end{equation}
where $k$ is a learnable scalar balancing the internal representation and external supervision.

NTI's residual shortcut reframes absolute prediction into a simple transformation,  analogous to the shortcuts in ResNet~\citep{he2016deep}, thus enabling gradients to bypass unstable backbone states and directly optimize the transfer layer.
To analyze this from a gradient perspective, let $\widehat{\bm{p}}_{n+2} = \operatorname{softmax}(\operatorname{logits}_2)$ denote the predicted probability distribution, and let $\bm{e}_{n+2}$ be the one-hot vector for the true token $u_{n+2}$. The gradient of the loss with respect to the transfer matrix is then:
\begin{equation}
\frac{\partial {\ell }^{\left( 2\right) }(\mathcal{D})}{\partial {\bm{W}}^{T}} = {\left( {\bm{W}}_{o}\left( {\bm{h}}_{n} + k{\left( {\bm{W}}_{t}\right) }_{\left(  : ,{u}_{n + 1}\right) }\right) \right) }^{\top }\left( {{\widehat{\bm{p}}}_{n + 2} - {\bm{e}}_{n + 2}}\right),
\end{equation}
thus preserving the informativeness of supervision even when the predicted next-step hidden state is corrupted by noise, thereby enhancing stability and accuracy in structural modeling.

\subsection{Transformer-Based Transfer Layer}

To overcome the limitations of linear mappings, we replace the linear transfer layer with a Transformer-based transfer layer. The input to this layer is the hidden representation \( \bm{h}_n \in \mathbb{R}^d \) produced by the backbone at the next-step position. Unlike linear layers that treat each dimension independently, the Transformer leverages self-attention to model dependencies across dimensions, allowing each component of \( \bm{h}_n \) to interact and integrate information from all others.

This dynamic interaction significantly enhances the expressiveness of the transfer layer, enabling more precise modeling of multi-hop relations in the underlying graph structure. Consequently, it achieves more accurate modeling of the transition from one-hop to multi-hop representations in Multi-Token Prediction tasks.

The overall architecture of the enhanced transfer layer, combining Next-Token Injection and Transformer-based transfer layer, is illustrated in Figure~\ref{fig:fig_NTI_transfer_arch}.

\section{Empirical Evaluation on Graph Planning}
\label{sec:experiments}

\subsection{Overall Accuracy of Different Models on Path Planning}
\label{5.1}

We evaluate model performance on randomly generated directed acyclic graphs (DAGs) by measuring prediction accuracy on test paths. To analyze performance under varying planning difficulties, test paths are categorized into degree-0/1/2/3 classes according to their reachability in the observation graph $\mathcal{G}_{\mathrm{obs}}$, as defined in Section~\ref{preliminaries}. 

\paragraph{Experimental Setup.}
For each trial, we generate a random DAG with $n=100$ nodes, where each potential edge $(i,j)$ for $i<j$ is included independently with probability $p=0.1$. For every reachable source–target pair $(s,t)$, 
	we randomly sample $m=20$ valid paths. To increase the number of test paths, 10\% of $(s,t)$ pairs are used for training and the remaining 90\% for testing, while all one-hop edges $(s,t)\in\mathcal{E}$ are always included in the training set as direct paths ``$s\ t\ s\ t\backslash n$''.  
All models use 120-dimensional embeddings and adopt a 1-layer, 1-head Transformer as the backbone. “NTI” denotes models with Next-Token Injection (Section~\ref{subsec:transfer_strategy}). The Transformer-based transfer layer uses the same hidden size as the backbone and varies in depth (1, 3, or 6 layers).

\paragraph{Metrics.}
We evaluate our models using three metrics. 
\textbf{Graph-level accuracy} is computed by first averaging the path-level accuracy within each graph, and then averaging across all graphs. 
\textbf{Standard error} is calculated by dividing the standard deviation of graph-level accuracies by the square root of the total number of graphs. 
\textbf{Path-level accuracy} is the average accuracy over all test paths.

\begin{table}[t]
	\caption{\textbf{Path prediction accuracy (\%) on degree-0/1/2/3 paths in 100-node DAGs.}
		Metrics include graph-level accuracy (with $\pm$ standard error) 
		and path-level accuracy. Results for degree-0/1 are averaged over 50 graphs; degree-2/3 over 200 graphs.}
	\label{tab:dag-eval-merged}
	\centering
	\fontsize{8}{9.5}\selectfont
	\resizebox{\linewidth}{!}{
		\begin{tabular}{lccccc}
			\toprule
			\multirow{2}{*}{\textsc{Model}}
			& \textsc{Degree-0} & \textsc{Degree-1} & \textsc{Degree-2} & \textsc{Degree-3} & \textsc{Overall} \\
			& Graph $\pm$ / Path & Graph $\pm$ / Path & Graph $\pm$ / Path & Graph $\pm$ / Path & Path \\
			\midrule
			1-Token (baseline)~\citep{wang2024alpine}
			& 92.58{\tiny $\pm$0.18} / 92.56
			& 86.57{\tiny $\pm$0.24} / 86.60
			& 63.80{\tiny $\pm$0.65} / 64.34
			& 30.76{\tiny $\pm$1.30} / 33.25
			& 89.31 \\
			\midrule
			2-Token (Meta's)~\citep{gloeckle2024better}
			& 92.03{\tiny $\pm$0.19} / 92.00
			& 85.54{\tiny $\pm$0.27} / 85.60
			& 66.37{\tiny $\pm$0.56} / 66.78
			& 36.09{\tiny $\pm$1.33} / 35.55
			& 88.65 \\
			2-Token (DeepSeek's)~\citep{liu2024deepseek}
			& 93.71{\tiny $\pm$0.18} / 93.62
			& 88.82{\tiny $\pm$0.22} / 88.79
			& 67.78{\tiny $\pm$0.58} / 68.48
			& 31.85{\tiny $\pm$1.34} / 34.36
			& 90.90 \\
			2-Token + NTI (linear transfer layer)
			& 94.09{\tiny $\pm$0.24} / 94.14
			& 90.00{\tiny $\pm$0.29} / 90.02
			& 69.51{\tiny $\pm$0.60} / 69.56
			& 39.25{\tiny $\pm$1.28} / 42.08
			& 91.74 \\
			2-Token + 1-layer Transformer
			& 93.87{\tiny $\pm$0.15} / 93.83
			& 87.84{\tiny $\pm$0.25} / 87.86
			& 68.78{\tiny $\pm$0.55} / 69.51
			& 37.11{\tiny $\pm$1.25} / 39.26
			& 90.68 \\
			2-Token + NTI + 1-layer
			& \textbf{96.43{\tiny $\pm$0.12} / 96.43}
			& 88.92{\tiny $\pm$0.21} / 88.94
			& 71.32{\tiny $\pm$0.51} / 71.52
			& 43.49{\tiny $\pm$1.17} / 44.37
			& \textbf{92.65} \\
			2-Token + NTI + 3-layer
			& 93.76{\tiny $\pm$0.16} / 93.70
			& 89.47{\tiny $\pm$0.24} / 89.39
			& 71.41{\tiny $\pm$0.55} / 71.60
			& 43.43{\tiny $\pm$1.30} / 44.26
			& 91.29 \\
			2-Token + NTI + 6-layer
			& 94.56{\tiny $\pm$0.17} / 94.50
			& \textbf{90.09{\tiny $\pm$0.23} / 90.10}
			& \textbf{72.56{\tiny $\pm$0.47} / 72.97}
			& \textbf{43.81{\tiny $\pm$1.23} / 45.70}
			& 92.07 \\
			\midrule
			3-Token (Meta's)~\citep{gloeckle2024better}
			& 90.76{\tiny $\pm$0.22} / 90.72
			& 83.37{\tiny $\pm$0.29} / 83.34
			& 62.17{\tiny $\pm$0.58} / 62.24
			& 34.63{\tiny $\pm$1.22} / 37.23
			& 86.90 \\
			3-Token (DeepSeek's)~\citep{liu2024deepseek}
			& \textbf{94.45{\tiny $\pm$0.16} / 94.37}
			& 89.42{\tiny $\pm$0.25} / 89.43
			& 66.21{\tiny $\pm$0.60} / 66.42
			& 30.22{\tiny $\pm$1.29} / 32.27
			& \textbf{91.54} \\
			3-Token + NTI (linear transfer layer)
			& 92.19{\tiny $\pm$0.15} / 92.17
			& 87.37{\tiny $\pm$0.20} / 87.39
			& 63.38{\tiny $\pm$0.56} / 64.06
			& 42.96{\tiny $\pm$1.36} / 46.28
			& 89.44 \\
			3-Token + 1-layer Transformer
			& 92.22{\tiny $\pm$0.15} / 92.16
			& 84.11{\tiny $\pm$0.22} / 84.15
			& 66.79{\tiny $\pm$0.47} / 67.22
			& 40.67{\tiny $\pm$1.24} / 40.43
			& 88.17 \\
			3-Token + NTI + 1-layer
			& 92.35{\tiny $\pm$0.16} / 92.31
			& 85.34{\tiny $\pm$0.24} / 85.37
			& 69.61{\tiny $\pm$0.49} / 70.10
			& 44.82{\tiny $\pm$1.31} / 46.10
			& 88.84 \\
			3-Token + NTI + 3-layer
			& 93.29{\tiny $\pm$0.13} / 93.25
			& 89.54{\tiny $\pm$0.16} / 89.52
			& 71.97{\tiny $\pm$0.47} / 72.39
			& \textbf{45.38{\tiny $\pm$1.18} / 47.25}
			& 91.12 \\
			3-Token + NTI + 6-layer
			& 93.55{\tiny $\pm$0.14} / 93.52
			& \textbf{89.67{\tiny $\pm$0.18} / 89.66}
			& \textbf{72.82{\tiny $\pm$0.49} / 73.09}
			& 45.18{\tiny $\pm$1.24} / 46.99
			& 91.34 \\
			\bottomrule
		\end{tabular}
	}
\end{table}

\paragraph{Results.}
As shown in Table~\ref{tab:dag-eval-merged}, the MTP models achieve notable improvements over the 1-Token Prediction (i.e., Next-Token Prediction) baseline on degree-2/3 test paths.
Compared to the MTP baseline, incorporating NTI and Transformer-based transfer layers consistently boosts performance across all degrees. 
The significant gains in accuracy for degree-2/3 tests in particular demonstrate the effectiveness of MTP in handling transitive reachability.

\begin{figure}[t]
  \centering
  \begin{minipage}[t]{0.45\textwidth}
    \vspace{0pt} 
    \centering
    \includegraphics[height=2.8cm]{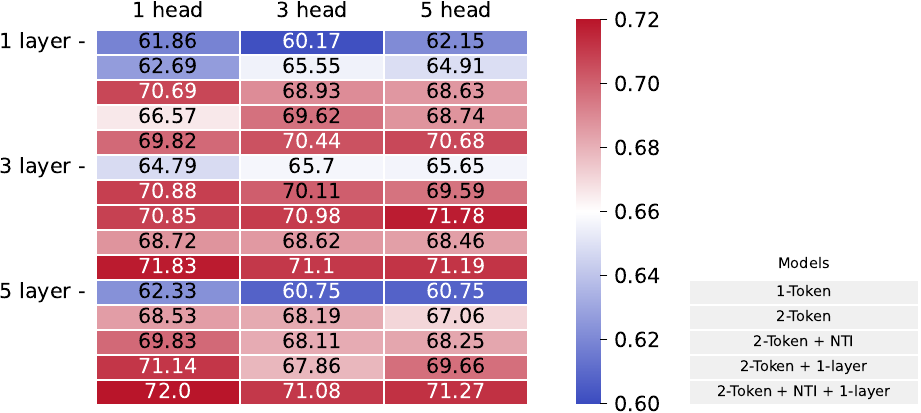}
    \caption{ 
    Degree-2 path graph-level accuracy (\%) for Transformer depth and number of heads, averaged over 10 fixed graphs.
    }
    \label{fig:backbone_config}
  \end{minipage}
  \begin{minipage}[t]{0.53\textwidth}
    \vspace{15pt} 
    \centering
    \captionof{table}{
    Degree-2 path graph-level accuracy (with $\pm$ standard error) for node counts and embedding sizes, averaged over 100 graphs.
    }
    \label{tab:node-embed-accuracy}
    \vspace{0em}
    \fontsize{8}{9.5}\selectfont
    \resizebox{\linewidth}{!}{
    \begin{tabular}{lcccc}
    \toprule
    \multirow{2}{*}{\textsc{Model}} 
    & \textsc{100-node} & \textsc{200-node} & \textsc{300-node} & \textsc{300-node} \\
    & \textsc{120-dim}  & \textsc{120-dim}  & \textsc{120-dim}  & \textsc{320-dim} \\
    \midrule
    1-Token & 63.00{\tiny $\pm$0.89} & 60.86{\tiny $\pm$0.82} & \textbf{56.16{\tiny $\pm$0.64}} & 55.92{\tiny $\pm$0.64} \\
    2-Token & 65.99{\tiny $\pm$0.88} & 62.43{\tiny $\pm$0.60} & 50.13{\tiny $\pm$0.69} & 57.47{\tiny $\pm$0.80} \\
    2-Token + NTI & 69.71{\tiny $\pm$0.85} & 63.84{\tiny $\pm$0.81} & 53.79{\tiny $\pm$0.60} & 65.32{\tiny $\pm$0.71} \\
    2-Token + 1-layer & 69.02{\tiny $\pm$0.69} & 61.30{\tiny $\pm$0.66} & 46.62{\tiny $\pm$0.52} & 60.51{\tiny $\pm$0.68} \\
    2-Token + NTI + 1-layer & \textbf{71.55{\tiny $\pm$0.73}} & \textbf{65.10{\tiny $\pm$0.62}} & 49.79{\tiny $\pm$0.54} & \textbf{65.14{\tiny $\pm$0.76}} \\
    \bottomrule
    \end{tabular}
    }
  \end{minipage}
  
\end{figure}

\paragraph{Effect of Backbone Architecture.}
We investigate the impact of backbone model complexity, including Transformer depth and number of attention heads. As shown in Figure~\ref{fig:backbone_config}, increasing backbone model complexity
	yields only marginal improvements. In contrast, the configuration combining 2-Token, NTI, and a Transformer-based transfer layer consistently yields stable accuracy gains.

\paragraph{Effect of Graph Size.}
We further evaluate scalability on graphs with 200 and 300 nodes, using a fixed embedding size of 120 (shown in Table~\ref{tab:node-embed-accuracy}). Results show that 2-Token performance degrades with increasing graph size. On 300-node graphs, increasing the embedding dimension to 320 enables the 2-Token model to outperform the 1-Token baseline. When further increasing the graph size and allocating a matching embedding dimension, the experimental results exhibit the same behavior. It is likely that when the number of nodes exceeds the embedding capacity, supervision signals from different tokens may conflict, limiting the model's ability to encode structural patterns.

\subsection{Weight Analysis of the Trained Model}
\label{sec:weights_analysis}

\begin{wrapfigure}{r}{0.50\textwidth}  
    \centering
    \vspace{-0.35cm} 
    \begin{subfigure}[t]{0.48\linewidth}
        \centering
        \includegraphics[height=2.5cm,width=\linewidth,keepaspectratio]{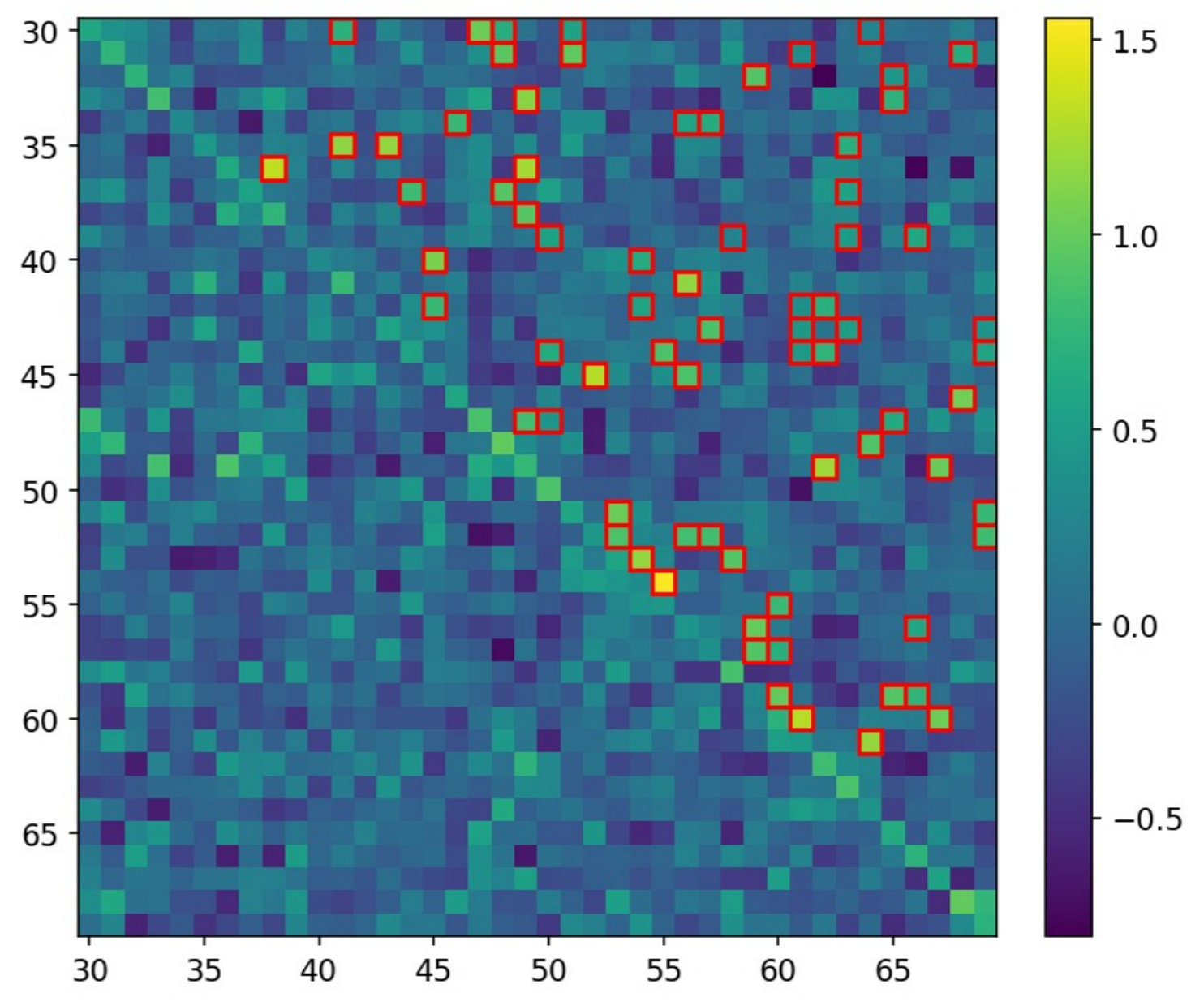}
        \caption{Without NTI}
        \label{fig:transfer-before}
    \end{subfigure}
    \hfill
    \begin{subfigure}[t]{0.48\linewidth}
        \centering
        \includegraphics[height=2.5cm,width=\linewidth,keepaspectratio]{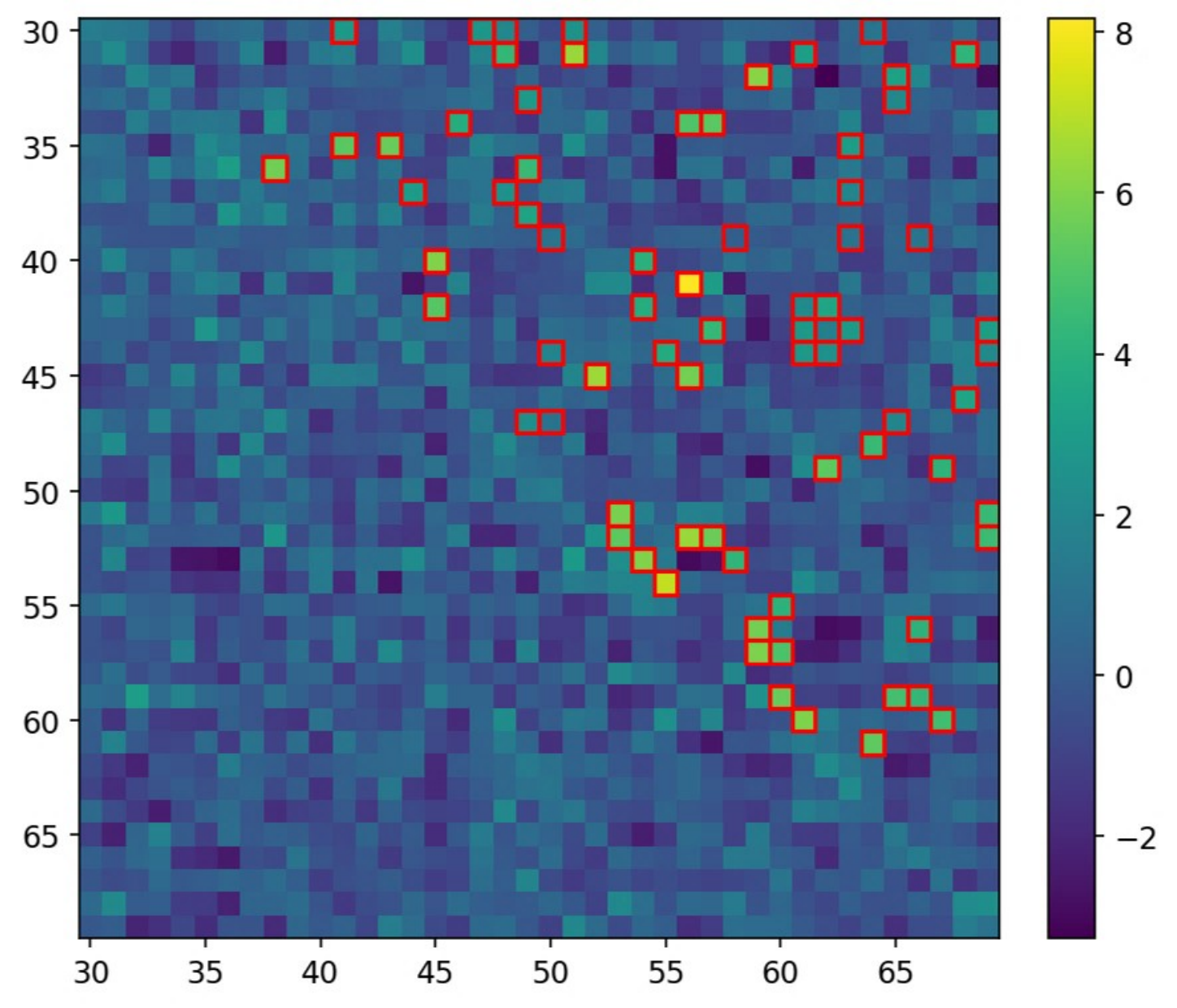}
        \caption{With NTI}
        \label{fig:transfer-after}
    \end{subfigure}

    \caption{
    \textbf{Visualization of the projected matrix on the 100-node graph without and with NTI.} 
Red boxes indicate ground-truth adjacency, showing only the central submatrix.}
    \label{fig:transfer-viz}
    \vspace{-0.5cm} 
\end{wrapfigure}

In a 100-node path-planning task, we project the transfer layer \(\bm{W}^T\) onto the node representations via \(\bm{W}_t\bm{W}^T\bm{W}_o\) to evaluate its ability to learn transition relations, where $\bm{W}_t$ and $\bm{W}_o$ denote the input embedding and output projection matrices, respectively. 
Figure~\ref{fig:transfer-viz} illustrates the visual projection and Table~\ref{tab:adj-weight} reports the average weights of the projection under true adjacency and non-adjacency entries.
The results clearly show that the transfer layer \(\bm{W}^T\) is learning the true adjacency and with NTI the learning effect is much better.

\begin{wraptable}{l}{0.40\textwidth} 
    \centering
    \vspace{-0.2cm}
    \caption{
    Average weights of adjacency and non-adjacency entries in the projected matrix and their gap.}
    \label{tab:adj-weight}
    \resizebox{\linewidth}{!}{ 
        \small
        \begin{tabular}{@{}lccc@{}}
            \toprule
            \textsc{Setting} & \textsc{Adj. Avg} & \textsc{Non-Adj. Avg} & \textsc{Gap} \\
            \midrule
            Without NTI & 0.82 & -0.01 & 0.83 \\
            With NTI    & 4.01 & -0.05 & 4.06 \\
            \bottomrule
        \end{tabular}
    }
\end{wraptable}

\begin{wrapfigure}{r}{0.58\textwidth} 
    \centering
    \vspace{-2.8cm} 
    \begin{subfigure}[t]{0.48\linewidth}
        \centering
        \includegraphics[height=3.5cm,keepaspectratio]{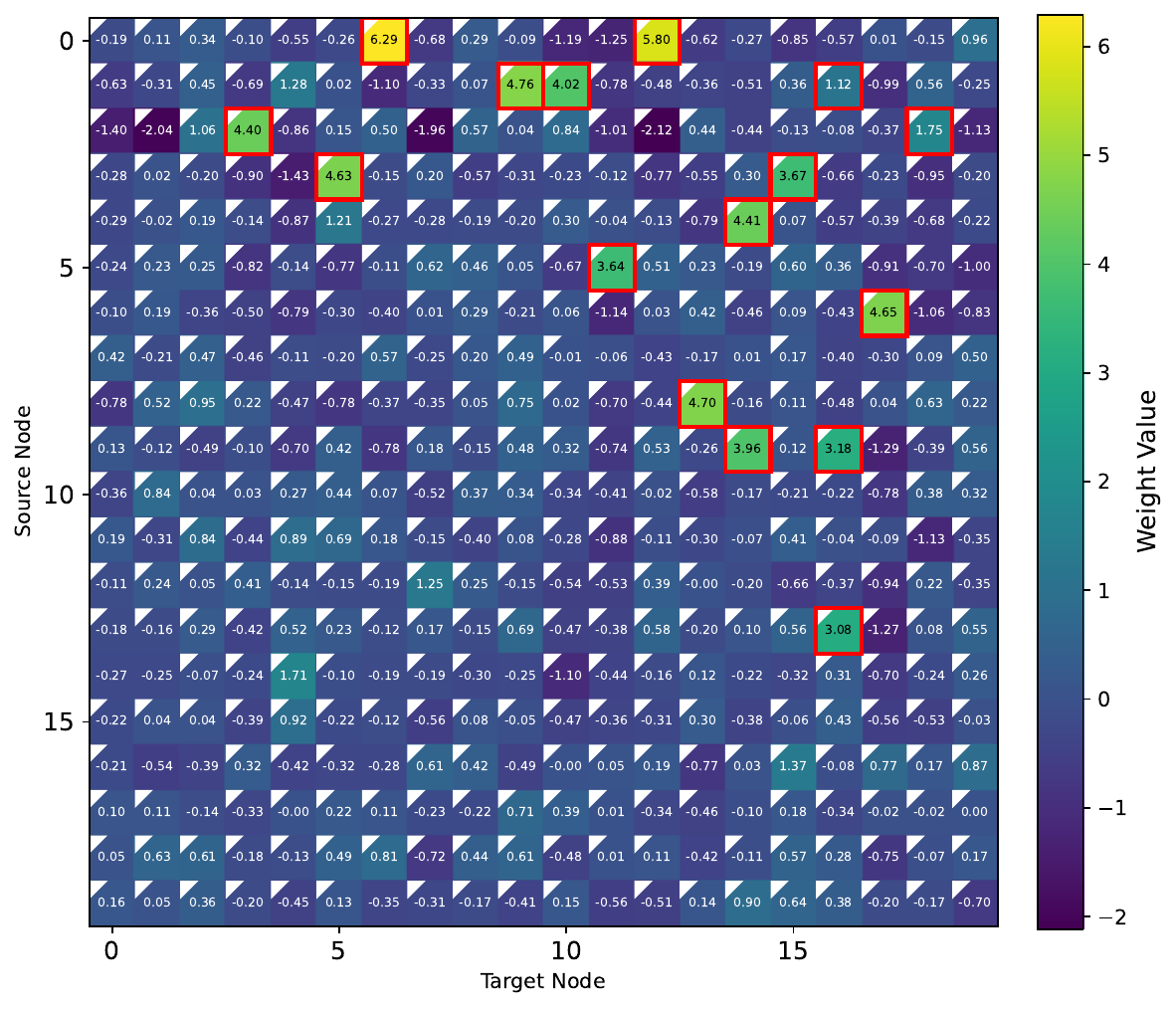}
        \caption{Learned $\bm{W}^M$}
        \label{fig:wm-heatmap-100-sub}
    \end{subfigure}
    \hfill
    \begin{subfigure}[t]{0.48\linewidth}
        \centering
        \includegraphics[height=3.5cm,keepaspectratio]{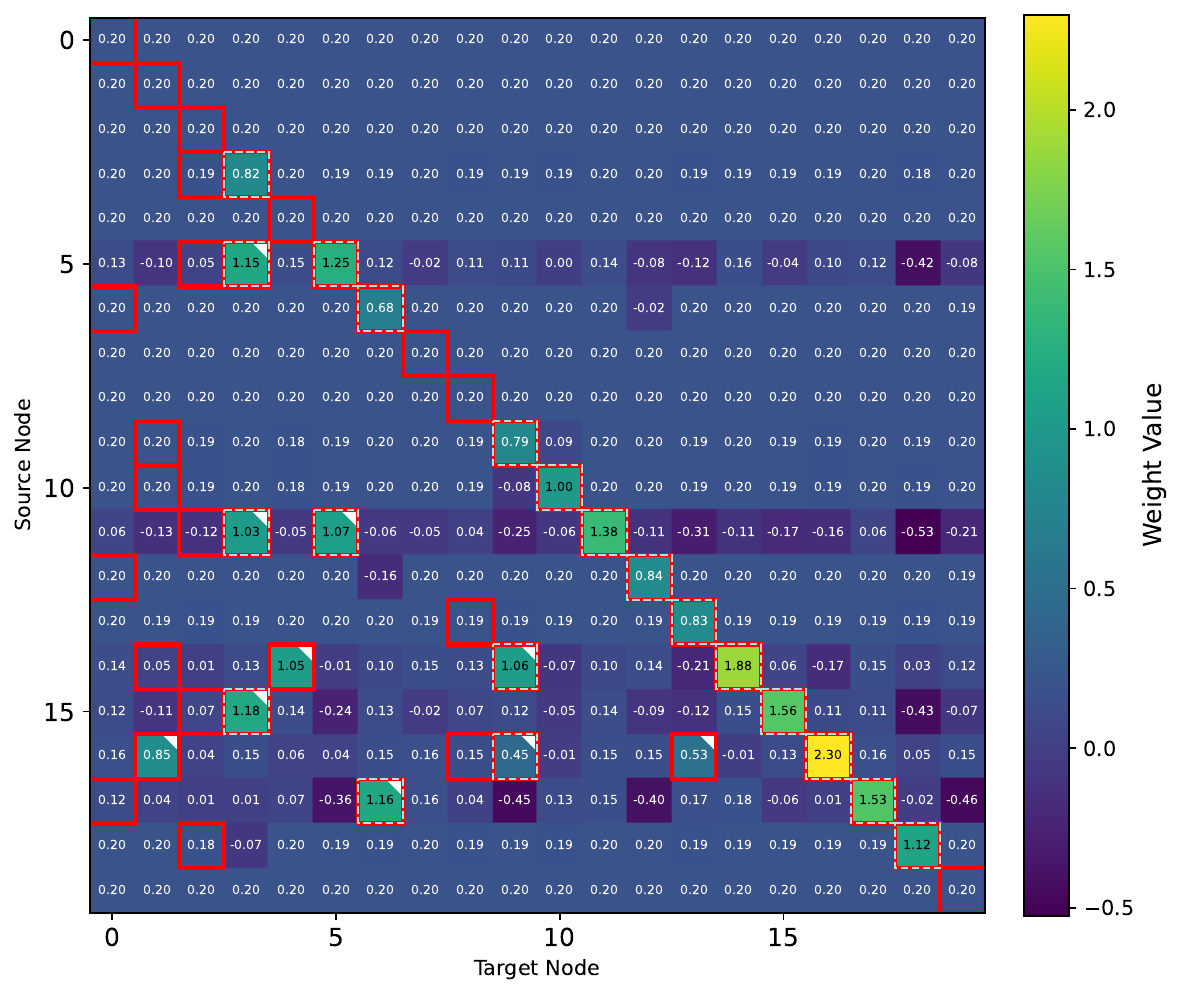}
        \caption{Learned $\bm{W}^V$}
        \label{fig:wv-heatmap-100-sub}
    \end{subfigure}

    \caption{\textbf{Weight analysis on the first 20 nodes of the 100-node graph using a simplified model with fixed transfer layer $\bm{W}^T$.} 
Red boxes indicate true adjacency and reachability, light dashed boxes show observed reachability, and white triangles mark theoretically learnable entries under $\ell^{(2)}(\mathcal{D})$.}

    \label{fig:wm-wv-combined-100-sub}
    \vspace{-0.4cm} 
\end{wrapfigure}

In the simplified 100-node model, with the transfer layer $\bm{W}^T$ fixed to the ground-truth adjacency matrix, we visualize the first 20 nodes of the learned weight matrices $\bm{W}^M$ and $\bm{W}^V$, as shown in Figures~\ref{fig:wm-heatmap-100-sub} and~\ref{fig:wv-heatmap-100-sub}. In $\bm{W}^M$, this region fully corresponds to relations that are theoretically learnable under $\ell^{(2)}(\mathcal{D})$, with some weights slightly amplified, though still far less pronounced than those reinforced by $\ell^{(1)}(\mathcal{D})$. In $\bm{W}^V$, a subset of relations is captured by $\ell^{(2)}(\mathcal{D})$, with their weights clearly standing out above the background (Table~\ref{tab:wm-wv-weights}).

The 2nd-step prediction encourages $\bm{W}^M$ to assign small positive weights to certain spurious adjacency relations, introducing mild noise. 
In contrast, $\bm{W}^V$ significantly strengthens correct but unobserved reachability relations, with weights clearly standing out from the background nonreachable entities. 
These enhanced reachability signals are concentrated near the diagonal in key positions, making them substantially more influential during the next-step prediction process.
This indicates that the 2nd-step prediction indeed enables the backbone model to learn transitive reachability beyond those learned by the next-token prediction.
Further visualizations and analyses of weights for the general 2-Token Prediction can be found in Appendix~\ref{appendix:weights}.

\begin{wraptable}{r}{0.5\linewidth}
\centering
\vspace{-0.45cm}
\caption{ Average weights of different entry types in $\bm{W}^M$ and $\bm{W}^V$ in the 100-node graph with a fixed transfer layer $\bm{W}^T$ in the simplified model.}
\label{tab:wm-wv-weights}
\resizebox{0.75\linewidth}{!}{
\begin{tabular}{clc}
\toprule
\textsc{Matrix} & \textsc{Type} & \textsc{Value} \\
\midrule
 & True adjacency & 1.90 \\
$\bm{W}^M$ & Theoretically learnable & 0.16 \\
 & Other entries & -0.01 \\
\midrule
 & Observed reachability & 0.63 \\
$\bm{W}^V$ & Theoretically learnable & 0.34 \\
 & Other entries & -0.01 \\
\bottomrule
\end{tabular}
}
\vspace{-0.4cm}
\end{wraptable}

\subsection{Evaluation on the Blocksworld Planning Task}
\label{blocksworld}

To further assess the practicality and generalization of our methods, we evaluate their performance on the classical Blocksworld planning benchmark. This task provides a structured, fixed-graph environment to test the models' ability in finding valid action sequences. 
Full details of the experimental setup, including the graph representation and dataset construction methodology, are provided in Appendix~\ref{app:blocksworld_details}.

\paragraph{Results.}
As shown in Table~\ref{tab:blocksworld-results}, our proposed enhancements to multi-token prediction consistently outperform the 1-token baseline and standard MTP architectures across various training data sizes. Notably, the combination of NTI and a multi-layer Transformer transfer layer (e.g., 2-Token + NTI + 6-layer) achieves the highest accuracy, demonstrating the effectiveness of our approach in a complex, classical planning environment.

\begin{table}[h]
\centering
\small
\caption{Path prediction accuracy (\%) on Blocksworld under varying training set sizes.}
\label{tab:blocksworld-results}
\resizebox{0.7\linewidth}{!}{
\begin{tabular}{lccccc}
\toprule
\textsc{Train Size (Number of Paths per Length)} & \textsc{100} & \textsc{200} & \textsc{300} & \textsc{400} & \textsc{500} \\
\midrule
1-Token (baseline) & 45.62 & 62.66 & 70.21 & 74.86 & 77.94 \\
\midrule
2-Token (Meta's)~\citep{gloeckle2024better} & 42.42 & 60.40 & 68.19 & 72.31 & 76.27 \\
2-Token (DeepSeek's)~\citep{liu2024deepseek} & 51.32 & 66.60 & 74.01 & 78.41 & 80.23 \\
2-Token + NTI (linear transfer layer) & 44.91 & 62.46 & 71.36 & 74.85 & 80.03 \\
2-Token + 1-layer Transformer & 51.25 & 66.51 & 75.84 & 77.11 & 79.32 \\
2-Token + NTI + 1-layer & \textbf{52.51} & 67.37 & \textbf{75.92} & \textbf{79.92} & 81.55 \\
2-Token + NTI + 3-layer & 52.01 & 66.73 & 74.86 & 79.56 & 83.30 \\
2-Token + NTI + 6-layer & 52.84 & \textbf{68.57} & 73.74 & 78.97 & \textbf{85.70} \\
\midrule
3-Token (Meta's)~\citep{gloeckle2024better} & 40.99 & 56.41 & 64.67 & 69.38 & 76.43 \\
3-Token (DeepSeek's)~\citep{liu2024deepseek} & 50.13 & 65.30 & 71.52 & 76.87 & 80.87 \\
3-Token + NTI (linear transfer layer) & 42.91 & 62.46 & 68.47 & 73.82 & 78.29 \\
3-Token + 1-layer Transformer & 49.98 & 64.95 & 72.63 & 76.83 & 78.37 \\
3-Token + NTI + 1-layer & \textbf{50.47} & \textbf{67.13} & \textbf{72.88} & 76.45 & 81.48 \\
3-Token + NTI + 3-layer & 49.84 & 66.57 & 71.70 & \textbf{77.47} & 80.40 \\
3-Token + NTI + 6-layer & 50.10 & 67.11 & 72.69 & 77.27 & \textbf{82.18} \\
\bottomrule
\end{tabular}}
\vspace{-0.2cm}
\end{table}

\section{Conclusion and Future Work}

The paper provides an in-depth analysis of how Multi-Token Prediction enables the autoregressive Transformer to learn
	transitive relation in graph path planning. 
Based on this, we propose two enhancement strategies: Next-Token Injection (NTI) and a Transformer-based Transfer Layer. Experiments on synthetic graphs and the Blocksworld planning task demonstrate that these methods significantly improve the model’s accuracy and stability in transitive planning tasks.

Our work opens several promising avenues for future research. 
One is to bridge the gap to real-world applications by effectively abstracting continuous and ambiguous tasks into discrete state representations for 
	multi-step prediction. 
Methodologically, our framework can be enhanced by extending the NTI mechanism to generate guiding signals in unsupervised settings, or by combining the transfer layer with explicit planning modules like chain-of-thought and backtracking search. Integrating these architectural improvements within a reinforcement learning paradigm presents another promising path toward creating more general and capable planning agents.

\newpage

\bibliography{iclr2026_conference}
\bibliographystyle{iclr2026_conference}

\newpage
\appendix
\section{Related Work}
\label{appendix:related-work}

\subsection{LLMs for Structured Planning and Reasoning}

Recent studies have examined the capacity of large language models (LLMs) to perform structured planning and reasoning. In planning domains, benchmarks such as CogEval~\citep{momennejad2023evaluating} and Blocksworld~\citep{valmeekam2023planning} highlight significant challenges: while humans achieve over 70\% success in Blocksworld, GPT-3 attains only 5\%, suggesting that LLMs struggle to capture underlying task structures. Nevertheless, LLMs exhibit promising behaviors in multi-step decision-making for autonomous agents~\citep{wang2024survey}, which can often be abstracted as path-finding over graphs. For example, HuggingGPT~\citep{shen2023hugginggpt} coordinates external APIs through dependency relations, naturally forming a graph-based planning problem. Our work adopts this abstraction but focuses on consistency and interpretability in multi-step prediction.

Beyond planning, researchers have explored the graph reasoning abilities of LLMs. Frameworks such as GPT4Graph~\citep{guo2023gpt4graph} and NLGraph~\citep{wang2023can} show that while LLMs can process graph-structured inputs, performance remains fragile and sensitive to spurious correlations, with GPT-4 reaching only around 50\% accuracy on shortest-path tasks. To improve reasoning, recent efforts augment LLMs with external modules such as GNN encoders~\citep{chai2023graphllm, tang2024graphgpt} or explicitly train them to imitate classical algorithms like BFS and DFS~\citep{luo2024graphinstruct}. Other work focuses on extracting structured task knowledge, e.g., distilling temporal and causal relations into compact representations~\citep{yuan2023distilling}. While these approaches improve empirical performance, they provide limited insights into why LLMs fail on more complex planning scenarios.

A complementary line of research examines the algorithmic foundations of LLM reasoning. Transformers have been shown to belong to the $\mathrm{TC}^0$ complexity class~\citep{merrill2023parallelism}, but techniques like chain-of-thought prompting~\citep{feng2023towards} and Tree of Thoughts search~\citep{yao2023tree} allow them to simulate more complex procedures sequentially. Parsel~\citep{zelikman2023parsel}, for instance, decomposes reasoning into structured subroutines, closely related to the multi-step prediction framework considered here. However, these studies largely overlook how the autoregressive training paradigm itself may impose fundamental limitations on consistent multi-step planning—a gap that our theoretical analysis seeks to address.

\subsection{Multi-Token Prediction (MTP)}

Traditional language models are trained with autoregressive next-token prediction (NTP), which inherently suffers from slow inference and discrepancies between training and inference~\citep{AL2023-cfg}. Moreover, NTP tends to overfit local transitions and struggles to capture long-range dependencies~\citep{bachmann2024pitfalls}.  
Multi-Token Prediction (MTP) has recently emerged as an alternative paradigm. Instead of predicting a single token at each step, MTP predicts multiple future tokens in parallel, often through several independent output heads attached to a shared Transformer backbone~\citep{gloeckle2024better, cai2024medusa}.

MTP offers two primary advantages. First, it enables self-speculative decoding, substantially accelerating inference. Approaches like Medusa~\citep{cai2024medusa} and Hydra~\citep{ankner2024hydra} verify multiple candidate tokens simultaneously, achieving up to $3\times$ speedup in low-batch scenarios. Second, by jointly predicting multiple steps, MTP encourages the model to capture longer-term dependencies and more global structures~\citep{gloeckle2024better, samragh2025your}, in contrast to NTP’s purely local supervision.

These benefits translate to strong empirical performance. In code generation, a 13B MTP-trained model outperforms its autoregressive counterpart by 12--17\% on HumanEval and MBPP~\citep{gloeckle2024better}. Beyond code, MTP has been applied to speech modeling and visual planning, demonstrating both substantial speedups and improved multi-step reasoning~\citep{wang2025vocalnet, zhang2025enhancing}.

From a theoretical perspective, ALPINE shows that autoregressive models face fundamental barriers in transitive planning problems~\citep{wang2024alpine}. By directly supervising multiple future tokens, MTP can mitigate exposure bias and better capture reachability structures essential for planning. In this work, we adopt MTP for training in graph-based path-finding tasks to explore its potential in overcoming the core limitations of NTP, while inference is still performed via NTP, as speedup is not our focus and we aim to study how multi-token supervision enhances learning.

\section{Simplified Transformer Setup}
\label{appendix:toy-analysis}

To theoretically analyze how the 2-Token Prediction objective helps the model learn structural information, we follow the analysis framework of~\citep{wang2024alpine}, starting from a standard Transformer model and gradually simplifying it into an analytically tractable form. We begin by reviewing the standard Transformer computations and then describe the simplifications introduced.

In the original model, the input tokens $x_1, x_2, \ldots, x_N$ are mapped into embedding vectors $\bm{X} \in \mathbb{R}^{N \times d}$, which are then processed through a stack of Transformer blocks. Each block consists of a multi-head self-attention (MHA) module and a feed-forward network (FFN). The attention mechanism is defined as:
\begin{equation}
\mathrm{Attention}(\bm{Q}, \bm{K}, \bm{V}) = \mathrm{softmax}\left( \frac{\bm{Q} \bm{K}^\top}{\sqrt{d_k}} \right) \bm{V},
\end{equation}
where $\bm{Q}, \bm{K}, \bm{V} \in \mathbb{R}^{N \times d_k}$ are the query, key, and value matrices respectively, $d_k = d / H$ is the per-head embedding dimension, and $H$ is the number of attention heads. For input $\bm{X} \in \mathbb{R}^{N \times d}$, the $i$-th head computes: \(\bm{Q}_i = \bm{X} \bm{W}_i^Q, \quad \bm{K}_i = \bm{X} \bm{W}_i^K, \quad \bm{V}_i = \bm{X} \bm{W}_i^V,\) with learnable parameters $\bm{W}_i^Q, \bm{W}_i^K, \bm{W}_i^V \in \mathbb{R}^{d \times d_k}$. The multi-head attention output is the concatenation of all heads: \(\mathrm{MHA}(\bm{X}) = \mathrm{Concat}_{i=1}^H \left( \mathrm{Attention}(\bm{Q}_i, \bm{K}_i, \bm{V}_i) \right).\)

The feed-forward network is a two-layer MLP:
\begin{equation}
\mathrm{FFN}(\bm{X}) = \max(\mathbf{0}, \bm{X} \bm{W}_1 + \mathbf{1} \bm{b}_1^\top) \bm{W}_2 + \mathbf{1} \bm{b}_2^\top,
\end{equation}
where $\bm{W}_1 \in \mathbb{R}^{d \times 4d}$, $\bm{W}_2 \in \mathbb{R}^{4d \times d}$, $\bm{b}_1 \in \mathbb{R}^{4d}$, $\bm{b}_2 \in \mathbb{R}^{d}$, $\mathbf{1} \in \mathbb{R}^{N \times 1}$ is the all-one vector for broadcasting biases, and $\mathbf{0} \in \mathbb{R}^{N \times 4d}$ is the zero matrix used in the ReLU activation \citep{glorot2011deep}.

Each Transformer block applies residual connections and layer normalizations:
\begin{equation}
\mathrm{Transformer}(\bm{X}) = \mathrm{FFN}(\mathrm{LN}_2(\mathrm{MHA}(\mathrm{LN}_1(\bm{X})) + \bm{X})) + \mathrm{MHA}(\mathrm{LN}_1(\bm{X})) + \bm{X}.
\end{equation}

\begin{figure}[htbp]
  \centering
  \begin{subfigure}[b]{0.48\textwidth}
    \centering
    \includegraphics[height=3.5cm]{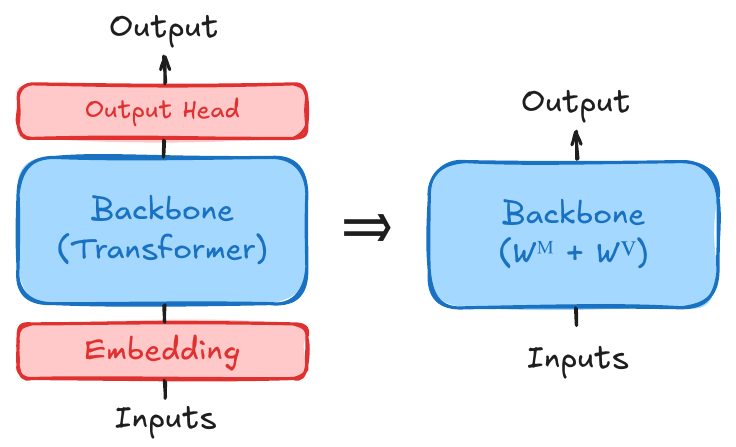}
    \caption{Simplified 1-Token Prediction model}
    \label{fig:1tp_model}
  \end{subfigure}
  \hfill
  \begin{subfigure}[b]{0.48\textwidth}
    \raggedright
    \includegraphics[height=3.5cm]{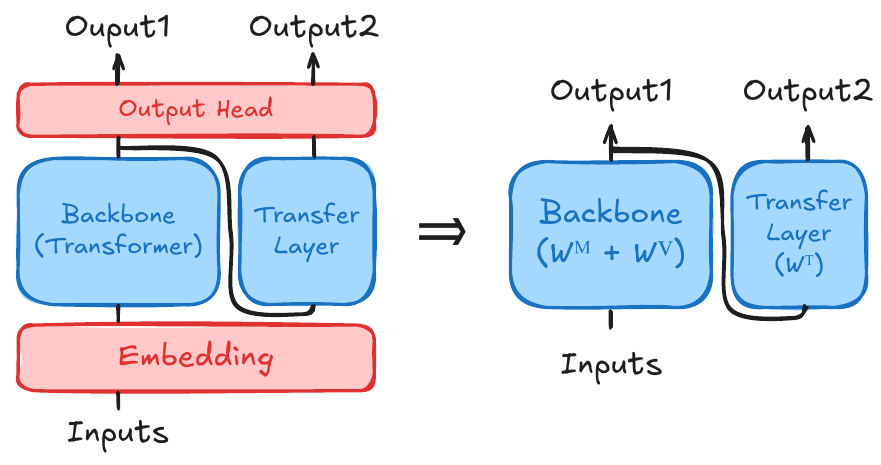}
    \caption{Simplified 2-Token Prediction model}
    \label{fig:2tp_model}
  \end{subfigure}
  \caption{Illustration for the simplified model architectures.}
  \label{fig:toy_models}
\end{figure}

\begin{figure}[t]
    \centering
    \begin{minipage}[b]{0.48\textwidth}
        \centering
        \includegraphics[height=3cm]{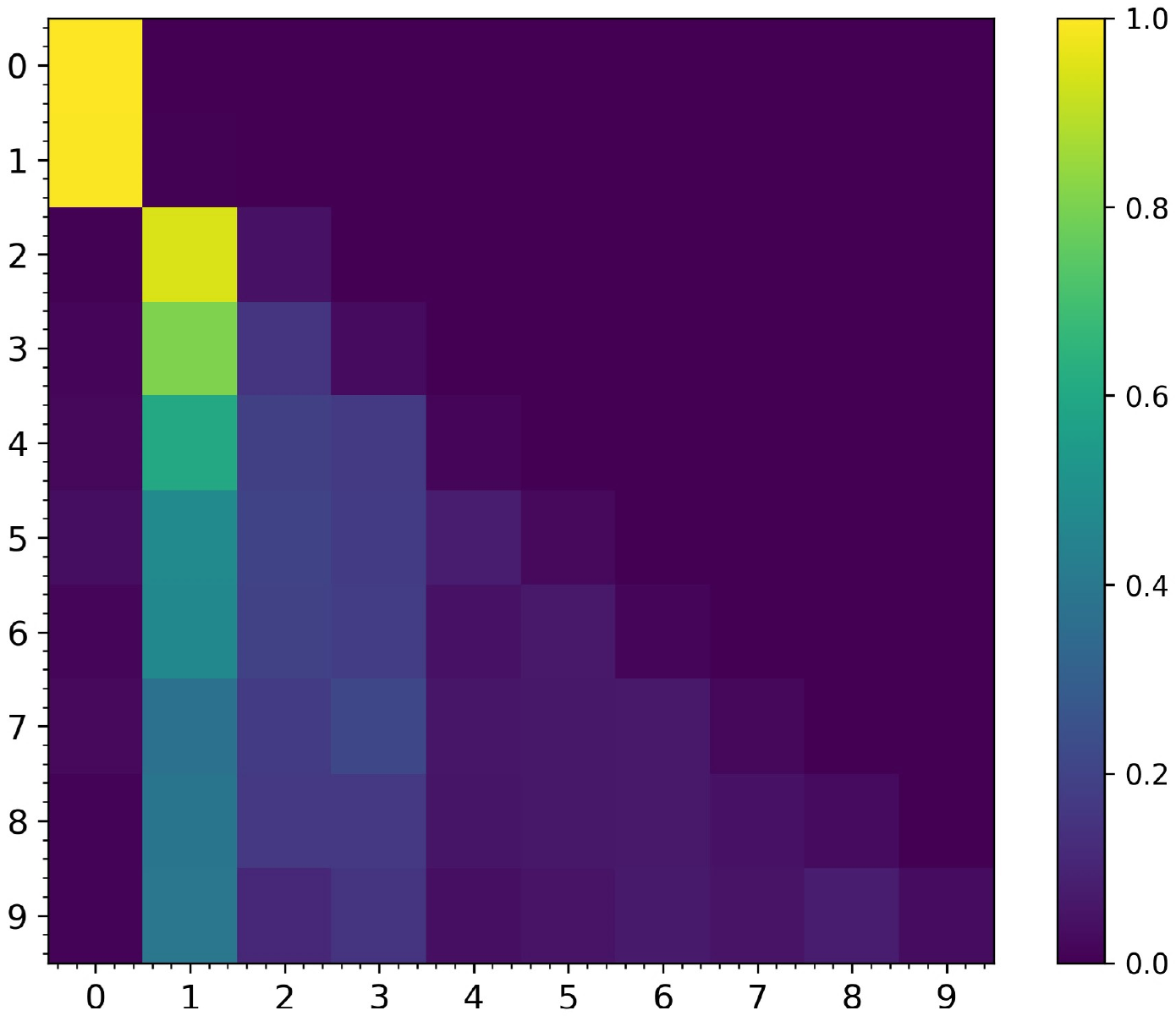}
        \caption*{(a) 1-Token Prediction model}
    \end{minipage}
    \hfill
    \begin{minipage}[b]{0.48\textwidth}
        \centering
        \includegraphics[height=3cm]{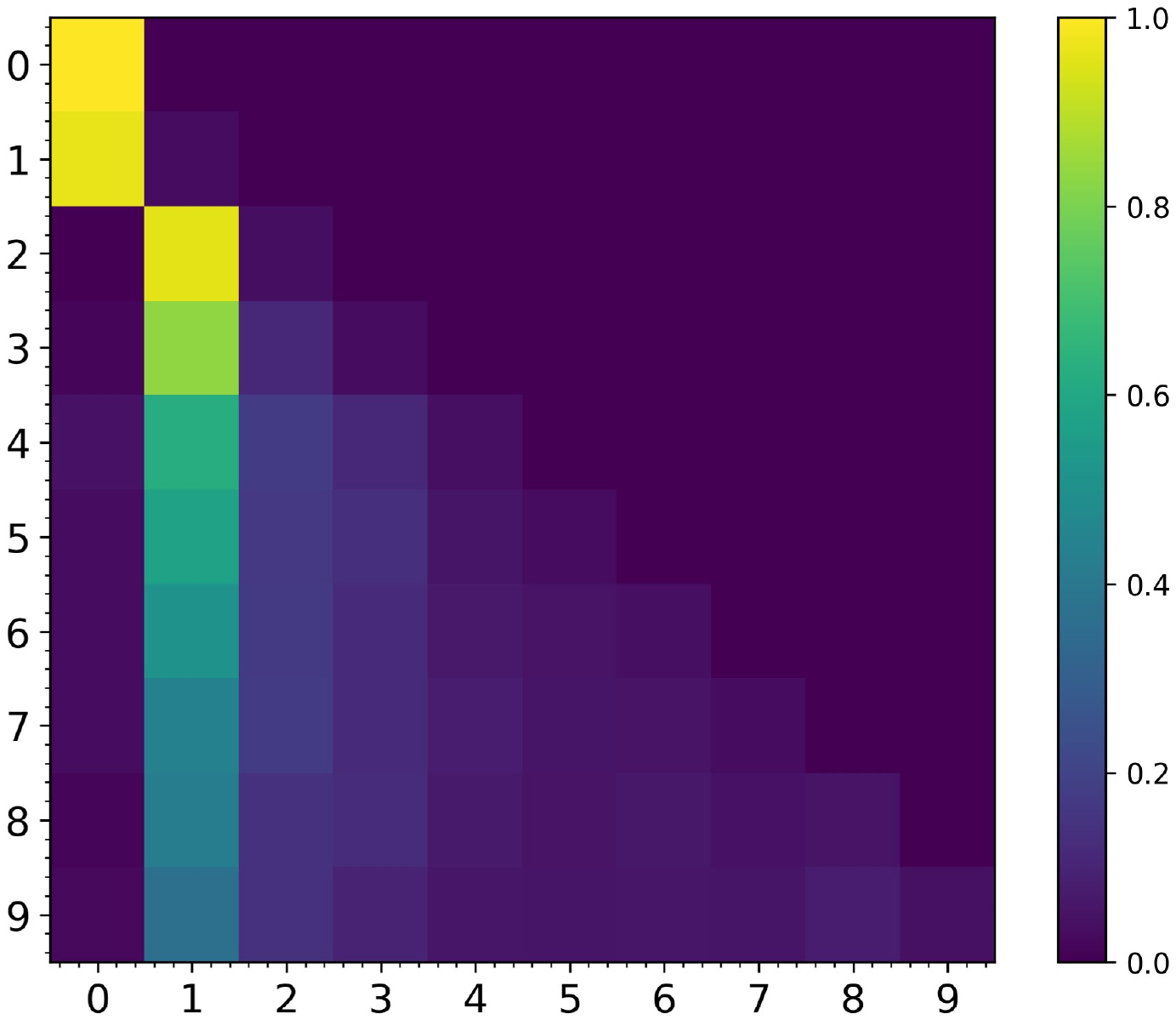}
        \caption*{(b) 2-Token Prediction model}
    \end{minipage}
    \caption{Visualization of attention matrices for (a) 1-Token Prediction and (b) 2-Token Prediction Transformer models.}
    \label{fig:toy_attention}
\end{figure}

To enable tractable visualization and analysis, we structurally simplify the architecture. First, we retain only a single-layer, single-head self-attention module. The attention matrix $\mathrm{softmax}\left(\frac{\bm{Q} \bm{K}^\top}{\sqrt{d_k}}\right)$ is manually set to a one-hot matrix in which the second column is filled with ones and all other entries are zero, i.e.,
\begin{equation}
\mathrm{softmax}\left(\frac{\bm{Q} \bm{K}^\top}{\sqrt{d_k}}\right) = 
\begin{bmatrix}
0 & 1 & 0 & \cdots & 0 \\
0 & 1 & 0 & \cdots & 0 \\
\vdots & \vdots & \vdots & \ddots & \vdots \\
0 & 1 & 0 & \cdots & 0 \\
\end{bmatrix} \in \mathbb{R}^{n \times n}.
\end{equation}
This setting simulates the attention matrix learned by the model after training on the task, where each token attends exclusively to the target node (i.e., the second token in the sequence). In the actual path-planning task with 100 nodes, the relevant results are shown in Figure~\ref{fig:toy_attention}. These results are obtained by analyzing the attention mechanism of a single-layer single-head Transformer model, presenting the averaged attention matrix computed over the test dataset. Each row \(n\) of the matrix corresponds to the attention distribution vector when predicting next token.

We remove all positional encodings by setting $\bm{W}_p = 0$, eliminate all layer normalization operations, and replace the two-layer FFN with a single linear transformation:
\begin{equation}
\mathrm{FFN}(\bm{X}) = \bm{X} \bm{W}^M,
\end{equation}
the forward propagation becomes an additive form. The resulting Transformer block is:
\begin{equation}
\mathrm{Transformer}(\bm{X}) = \mathrm{FFN}(\bm{X}) + \mathrm{MHA}(\bm{X}).
\end{equation}

To further reduce complexity, we set both the token embedding matrix $\bm{W}_t$ and the output projection matrix $\bm{W}_o$ to identity matrices, and assume that the embedding dimension equals the vocabulary size, i.e., $d = M$. This allows direct interpretation of logits in the vocabulary space.

Finally, in the 2-Token Prediction setting, we introduce a transfer matrix $\bm{W}^T \in \mathbb{R}^{M \times M}$ after the logits layer. This transfer layer is used to map the predicted next-step logits to the logits for the following step. The two resulting model variants are shown in Figure~\ref{fig:toy_models}.

\section{Derivations and Proofs}
\label{appendix:gradient-analysis}

Let $N_{i,j,k'}$ denote the number of times in $\mathcal{D}$ that the following conditions are satisfied: 
(a) the current node is $i$; 
(b) the attention target is $j$; and 
(c) the token two steps ahead is $k'$. 
Let $N_{i,j} = \sum_{k'} N_{i,j,k'}$ denote the total count of such $(i, j)$ pairs. 

This leads to the following theorem:

\textbf{Theorem 1.}  
\textit{For any pair \((i, j)\) in dataset \(\mathcal{D}\) with \(N_{i,j} > 0\), let \(P^{\mathrm{data}}_{i,j}(k') = \frac{N_{i,j,k'}}{N_{i,j}}\) be the empirical probability of the second-next node \(k'\). The contribution of this pair to the gradient \(\frac{\partial \ell^{(2)}(\mathcal{D})}{\partial \bm{W}^T_{(d,k')}}\) is determined by the prediction error, for any \(d\) where \((\bm{W}^M_{(i,d)} + \bm{W}^V_{(j,d)}) > 0\):
(i) If \(\widehat{P}_{i,j}(k') < P^{\mathrm{data}}_{i,j}(k')\), the contribution is \textbf{negative}, promoting an \textbf{increase} in the weight \(\bm{W}^T_{(d,k')}\).
(ii) Conversely, if \(\widehat{P}_{i,j}(k') > P^{\mathrm{data}}_{i,j}(k')\), the contribution is \textbf{positive}, promoting a \textbf{decrease} in the weight.
The total gradient is the sum of these contributions over all pairs \((i, j)\) in \(\mathcal{D}\).}

\textbf{Theorem 2.}  

\textit{
For any pair \((i, j)\) in dataset \(\mathcal{D}\) with \(N_{i,j} > 0\), the contribution of each \((\text{current node } i, \text{second-step node } k')\) pair to the gradient \(\frac{\partial \ell^{(2)}(\mathcal{D})}{\partial \bm{W}^V_{(j,k)}}\) is determined by the prediction error, for any \(k\) where \(\bm{W}^T_{(k,k')} > 0\):  
(i) If \(\widehat{P}_{i,j}(k') < P^{\mathrm{data}}_{i,j}(k')\), the contribution is \textbf{negative}, promoting an \textbf{increase} in the weight \(\bm{W}^V_{(j,k)}\);  
(ii) Conversely, if \(\widehat{P}_{i,j}(k') > P^{\mathrm{data}}_{i,j}(k')\), the contribution is \textbf{positive}, promoting a \textbf{decrease} in the weight.  
The total gradient is the sum of contributions from all \((i, k')\) pairs.}

\textit{
For any pair \((i, j)\) in dataset \(\mathcal{D}\) with \(N_{i,j} > 0\), the contribution of each \((\text{target node } j, \text{second-step node } k')\) pair to the gradient \(\frac{\partial \ell^{(2)}(\mathcal{D})}{\partial \bm{W}^M_{(i,k)}}\) is determined by the prediction error, for any \(k\) where \(\bm{W}^T_{(k,k')} > 0\):  
(i) If \(\widehat{P}_{i,j}(k') < P^{\mathrm{data}}_{i,j}(k')\), the contribution is \textbf{negative}, promoting an \textbf{increase} in the weight \(\bm{W}^M_{(i,k)}\);  
(ii) Conversely, if \(\widehat{P}_{i,j}(k') > P^{\mathrm{data}}_{i,j}(k')\), the contribution is \textbf{positive}, promoting a \textbf{decrease} in the weight.  
The total gradient is the sum of contributions from all \((j, k')\) pairs.  
}

\textit{Proof.}

According to the definition of cross-entropy loss and the predicted weight vectors in our simplified model, the total cross-entropy loss (involving matrices \(\bm{W}^M\), \(\bm{W}^V\), and \(\bm{W}^T\)) is given by
\begin{align*}
\ell^{(2)}(\mathcal{D}) 
&= - \sum_{u \in \mathcal{D}} \sum_{n=1}^{N-2} \sum_k \bm{U}_{(n+2, k)} \log \frac{ \exp\left( \left( \bm{W}^M \bm{W}^T \right)_{(u_n, k)} + \left( \bm{W}^V \bm{W}^T \right)_{(u_2, k)} \right) }{ \sum_{\ell} \exp\left( \left( \bm{W}^M \bm{W}^T \right)_{(u_n, \ell)} + \left( \bm{W}^V \bm{W}^T \right)_{(u_2, \ell)} \right) } \\
\text{Let} \quad 
\bm{A}_{(i,k)} &= \left( \bm{W}^M \bm{W}^T \right)_{(i,k)}, \quad 
\bm{B}_{(j,k)} = \left( \bm{W}^V \bm{W}^T \right)_{(j,k)} \\
\ell^{(2)}(\mathcal{D}) 
&= - \sum_{u \in \mathcal{D}} \sum_{n=1}^{N-2} \sum_k \bm{U}_{(n+2, k)} \sum_{i,j} \mathbb{I}[u_n = i, u_2 = j] 
\log \frac{ \exp( \bm{A}_{(i,k)} + \bm{B}_{(j,k)} ) }{ \sum_{\ell} \exp( \bm{A}_{(i,\ell)} + \bm{B}_{(j,\ell)} ) } \\
&= - \sum_{i,j,k'} N_{i,j,k'} \log \frac{ \exp( \bm{A}_{(i,k')} + \bm{B}_{(j,k')} ) }{ \sum_{\ell} \exp( \bm{A}_{(i,\ell)} + \bm{B}_{(j,\ell)} ) } \\
&= - \sum_{i,j,k'} N_{i,j,k'} ( \bm{A}_{(i,k')} + \bm{B}_{(j,k')} ) 
+ \sum_{i,j} N_{i,j} \log \left( \sum_{\ell} \exp( \bm{A}_{(i,\ell)} + \bm{B}_{(j,\ell)} ) \right) \\
&= - \sum_{i,j,k'} N_{i,j,k'} \left( \left( \bm{W}^M \bm{W}^T \right)_{(i,k')} + \left( \bm{W}^V \bm{W}^T \right)_{(j,k')} \right) \\
&\quad + \sum_{i,j} N_{i,j} \log \left( \sum_{\ell} \exp\left( \left( \bm{W}^M \bm{W}^T \right)_{(i,\ell)} + \left( \bm{W}^V \bm{W}^T \right)_{(j,\ell)} \right) \right).
\end{align*}

We define \(\hat{P}_{i,j}(k')\) as the softmax probability—under current model parameters—that, given current node \(i\) and target node \(j\), the model predicts node \(k'\) as the token at step \(n+2\):
\[
\widehat{P}_{i,j}(k') = \frac{\exp\left( \bm{A}_{(i,k')} + \bm{B}_{(j,k')} \right)}{\sum_{\ell} \exp\left( \bm{A}_{(i,\ell)} + \bm{B}_{(j,\ell)} \right)}.
\]

Then we have that the total gradient is the sum of contributions from all pairs \((i, j)\):
\begin{equation} \label{eq:gradient-WT}
\frac{\partial \ell^{(2)}(\mathcal{D})}{\partial \bm{W}^T_{(d,k')}} = 
\sum_{i,j} \underbrace{\left[ \left( \widehat{P}_{i,j}(k') - P^{\mathrm{data}}_{i,j}(k') \right) \cdot N_{i,j} \cdot \left( \bm{W}^M_{(i,d)} + \bm{W}^V_{(j,d)} \right) \right]}_{\text{Contribution from pair \((i, j)\)}}.
\end{equation}
The proof of Theorem~\ref{thm:1} follows by analyzing the sign of each term---or contribution---in this summation. For any specific pair \((i, j)\) with \(N_{i,j} > 0\), the sign of its contribution is determined by the prediction error.

If \(\widehat{P}_{i,j}(k') < P^{\mathrm{data}}_{i,j}(k')\), the first factor in the term, \(\left( \widehat{P}_{i,j}(k') - P^{\mathrm{data}}_{i,j}(k') \right)\), is negative. Given the theorem's conditions that \(N_{i,j} > 0\) and \((\bm{W}^M_{(i,d)} + \bm{W}^V_{(j,d)}) > 0\), the entire term corresponding to this \((i, j)\) pair is therefore \textbf{negative}, which proves part (i) of the theorem. Conversely, if \(\widehat{P}_{i,j}(k') > P^{\mathrm{data}}_{i,j}(k')\), the first factor is positive, making the entire term for this \((i, j)\) pair \textbf{positive}. This proves part (ii) of the theorem.

The total gradient is the aggregation of all such positive and negative contributions from every pair \((i,j)\) in the dataset \(\mathcal{D}\), as stated in the final sentence of the theorem. This concludes the proof.

Similarly, the gradient with respect to $\bm{W}^M$ is derived as the sum of contributions from all applicable pairs \((j, k')\):
\begin{equation} \label{eq:gradient-WM}
\frac{\partial \ell^{(2)}(\mathcal{D})}{\partial \bm{W}^M_{(i,k)}} = 
\sum_{j,k'} \underbrace{\left[ \left( \widehat{P}_{i,j}(k') - P^{\mathrm{data}}_{i,j}(k') \right) \cdot N_{i,j} \cdot \bm{W}^T_{(k,k')} \right]}_{\text{Contribution from pair \((j,k')\)}}.
\end{equation}
The proof of Theorem 2 for the backbone parameter $\bm{W}^M$ follows from analyzing the sign of each term in this summation. For any term in this sum that corresponds to a context \((i, j)\) with \(N_{i,j} > 0\), its sign is determined by the prediction error for that context.

If \(\widehat{P}_{i,j}(k') < P^{\mathrm{data}}_{i,j}(k')\), the initial factor \(\left( \widehat{P}_{i,j}(k') - P^{\mathrm{data}}_{i,j}(k') \right)\) is negative. Given the theorem's conditions that \(N_{i,j} > 0\) and \(\bm{W}^T_{(k,k')} > 0\), the entire term corresponding to this \((j,k')\) pair is therefore \textbf{negative}, which proves part (i) of the theorem regarding the contribution to the gradient of $\bm{W}^M_{(i,k)}$. Conversely, if \(\widehat{P}_{i,j}(k') > P^{\mathrm{data}}_{i,j}(k')\), the initial factor is positive, making the entire term for this \((j,k')\) pair \textbf{positive}. This proves part (ii) of the theorem.

The total gradient for \(\bm{W}^M_{(i,k)}\) is the sum of all such positive and negative contributions over all applicable pairs \((j, k')\). As stated in the theorem, analogous reasoning holds for gradients w.r.t.\ $\bm{W}^V$. This concludes the proof.

\section{Further Weight Analysis of Simplified Model}
\label{appendix:weights}

Fixing the transfer layer $\bm{W}^T$ to the ground-truth adjacency, we train the simplified model on a 20-node graph.  

\begin{figure}[h]
    \centering
    \begin{subfigure}[t]{0.48\linewidth}
        \centering
        \includegraphics[height=5cm]{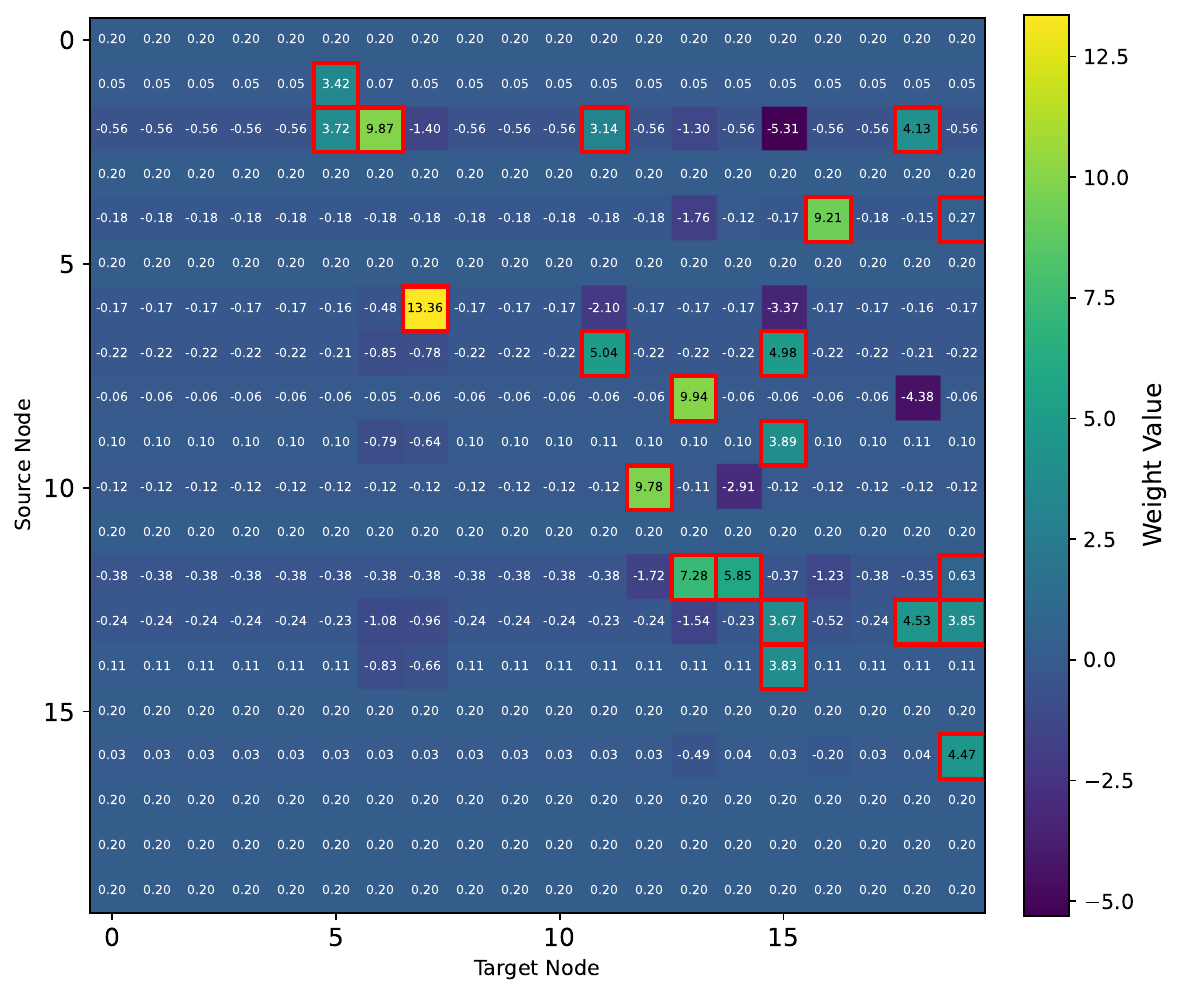}
        \caption{Learned $\bm{W}^M$}
        \label{fig:wm-heatmap_tot20}
    \end{subfigure}
    \hfill
    \begin{subfigure}[t]{0.48\linewidth}
        \centering
        \includegraphics[height=5cm]{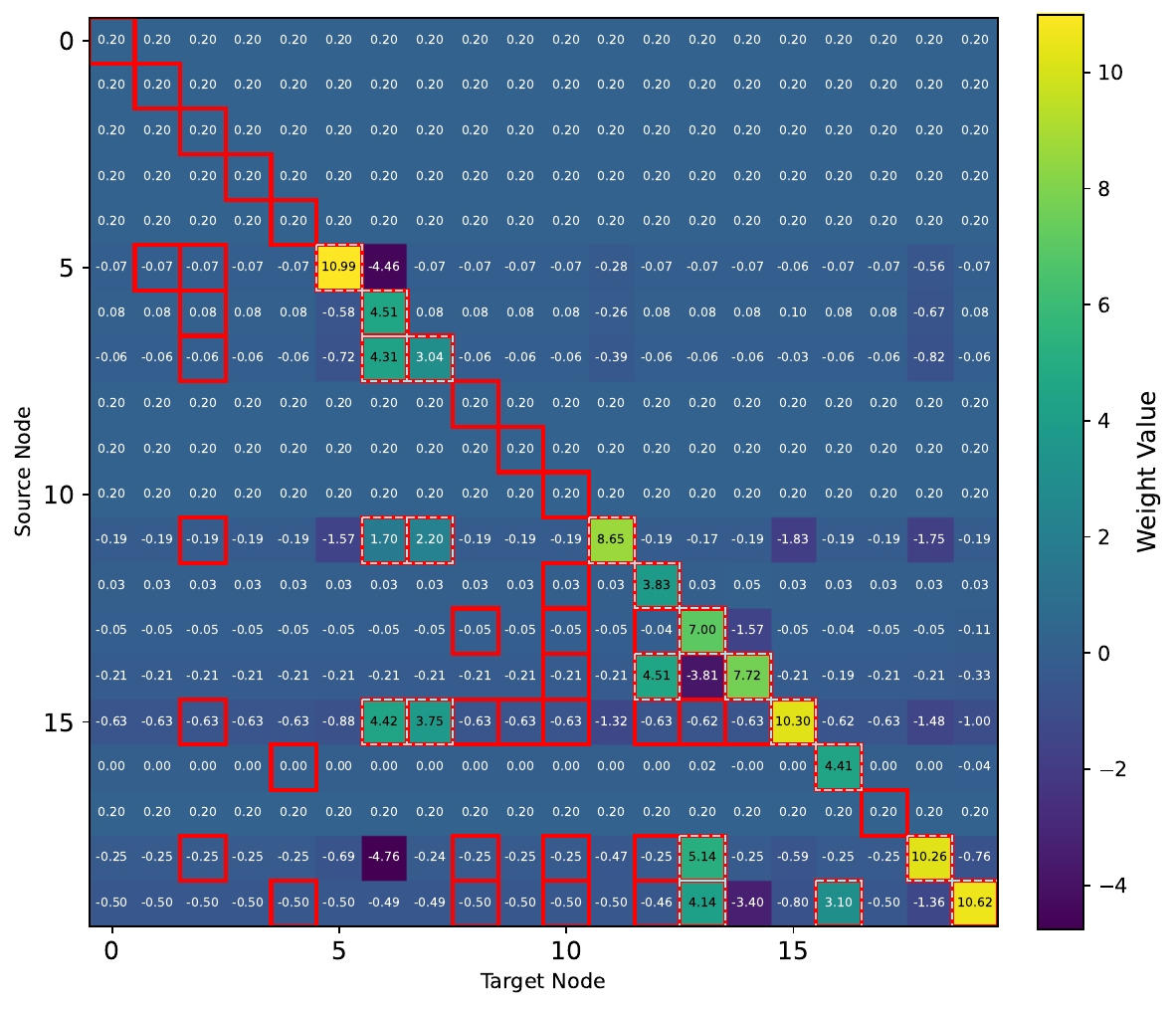}
        \caption{Learned $\bm{W}^V$}
        \label{fig:wv-heatmap_tot20}
    \end{subfigure}
    \caption{\textbf{Weight visualizations on a 20-node graph trained with 1-Token Prediction.} Red boxes highlight true adjacency in $\bm{W}^M$ (left) and true reachability in $\bm{W}^V$ (right). In the right plot, light dashed boxes indicate observed reachability.}
    \label{fig:wm-wv-combined_1}
\end{figure}

\begin{figure}[h]
    \centering
    \begin{subfigure}[t]{0.48\linewidth}
        \centering
        \includegraphics[height=5cm]{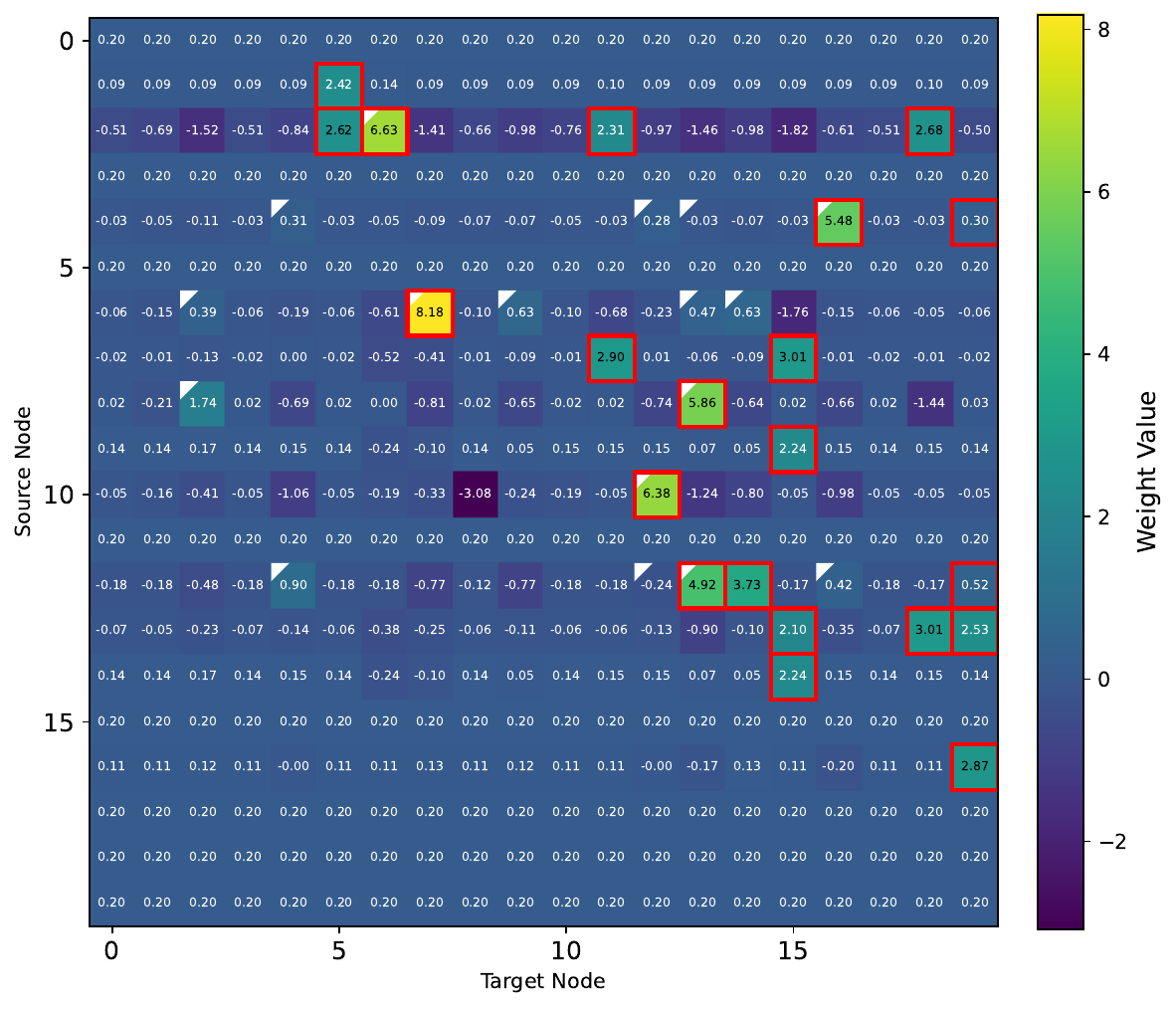}
        \caption{Learned $\bm{W}^M$}
        \label{fig:wm-heatmap}
    \end{subfigure}
    \hfill
    \begin{subfigure}[t]{0.48\linewidth}
        \centering
        \includegraphics[height=5cm]{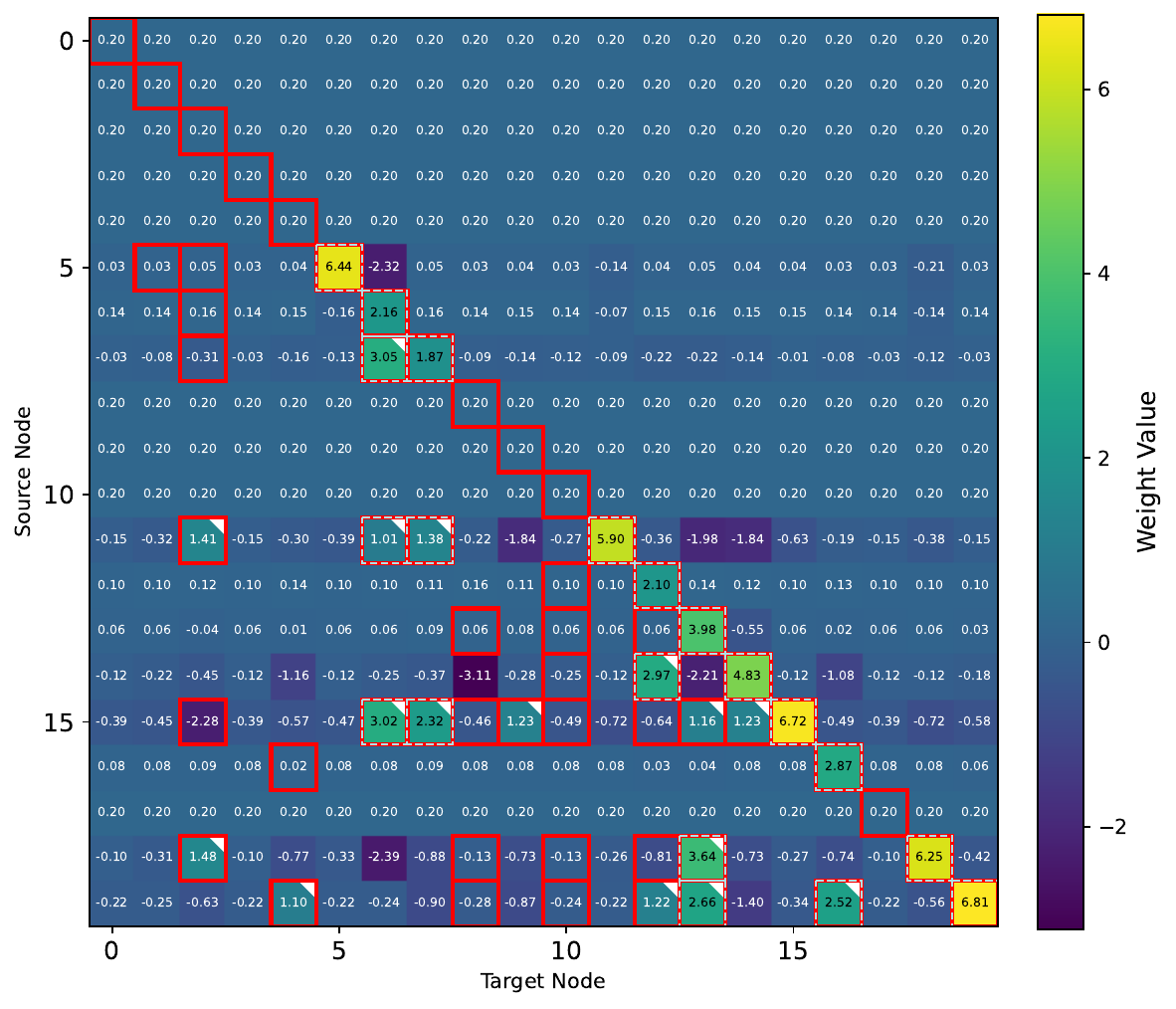}
        \caption{Learned $\bm{W}^V$}
        \label{fig:wv-heatmap}
    \end{subfigure}
    \caption{\textbf{Weight visualizations on a 20-node graph with fixed transfer layer $\bm{W}^T$ trained with 2-Token Prediction.} Red boxes highlight true adjacency in $\bm{W}^M$ (left) and true reachability in $\bm{W}^V$ (right). In the right plot, light dashed boxes indicate observed reachability. White triangles indicate theoretically learnable entries under $\ell^{(2)}(\mathcal{D})$.}
    \label{fig:wm-wv-combined}
\end{figure}

Figure~\ref{fig:wm-wv-combined_1} shows the learned weight matrices after training with 1-Token Prediction. As can be seen, the model learns the adjacency relations in $\bm{W}^M$ and the observed reachability relations in $\bm{W}^V$. Figure~\ref{fig:wm-wv-combined} shows the learning results of 2-Token Prediction, where the matrices contain entries that are theoretically learned under the $\ell^{(2)}(\mathcal{D})$ constraint. These entries have weights slightly higher than the background, but remain significantly weaker than those reinforced by $\ell^{(1)}(\mathcal{D})$. Ultimately, spurious adjacency relations are observed in $\bm{W}^M$, while $\bm{W}^V$ partially captures unobserved reachability relations.

To rigorously validate our theoretical analysis, Figure~\ref{fig:wm-wv-combined-100-sub} in the main text shows the training results of 2-Token Prediction on a 100-node graph when the transfer layer $\bm{W}^T$ is fixed to the ground-truth adjacency matrix. However, in the general 2-Token Prediction setting, the transfer layer $\bm{W}^T$ is learned, with its input being the backbone-predicted next-step logits rather than the true next-step one-hot vectors. As a result, it exhibits certain deviations from the ground-truth adjacency matrix.

\begin{figure}[h]
    \centering
    
    \begin{subfigure}[t]{0.48\linewidth}
        \centering
        \includegraphics[height=5cm]{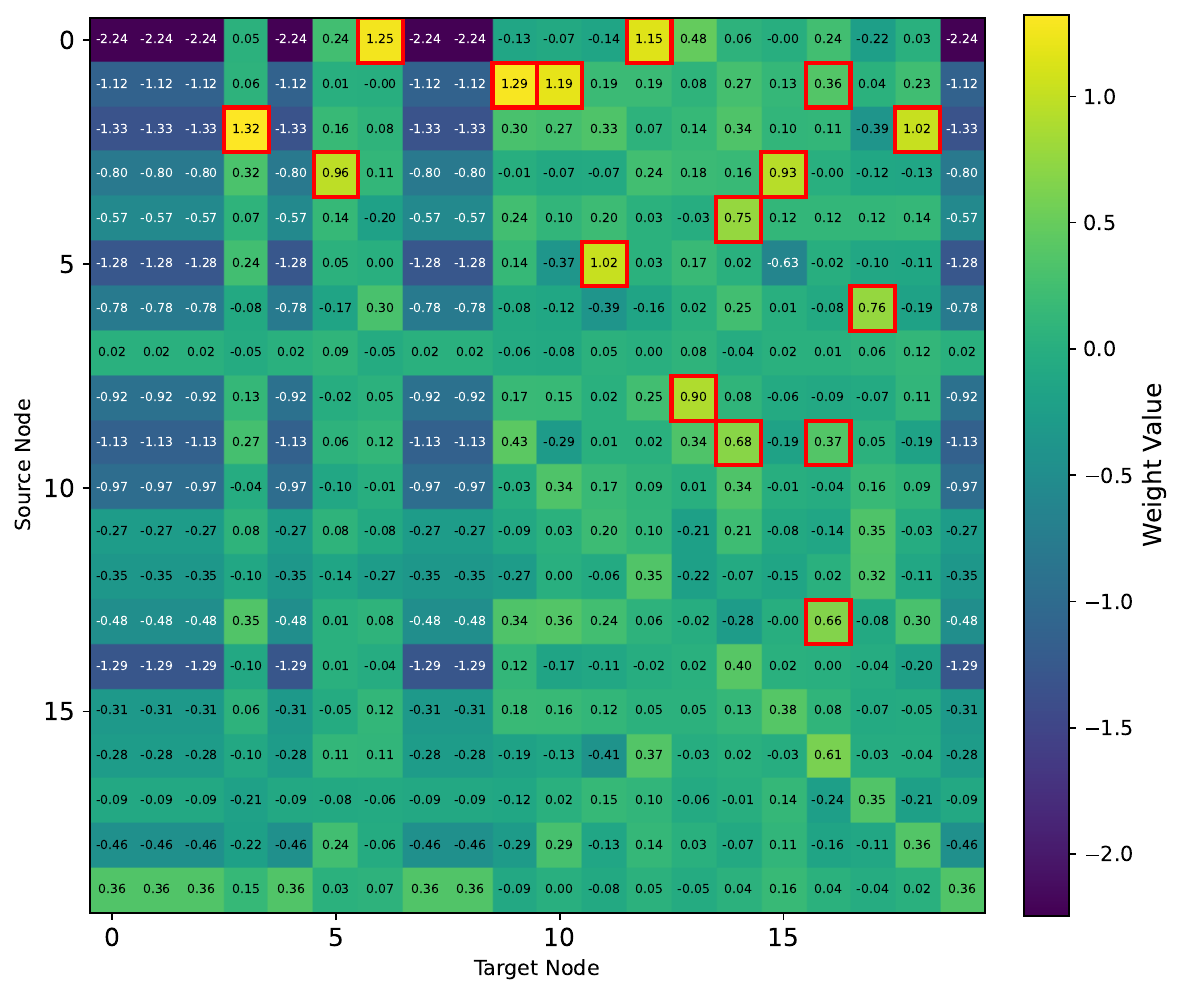} 
        \caption{Learned $\bm{W}^T$ with true adjacency}
        \label{fig:wt-heatmap-20}
    \end{subfigure}
    \vskip\baselineskip
    
    \begin{subfigure}[t]{0.48\linewidth}
        \centering
        \includegraphics[height=5cm]{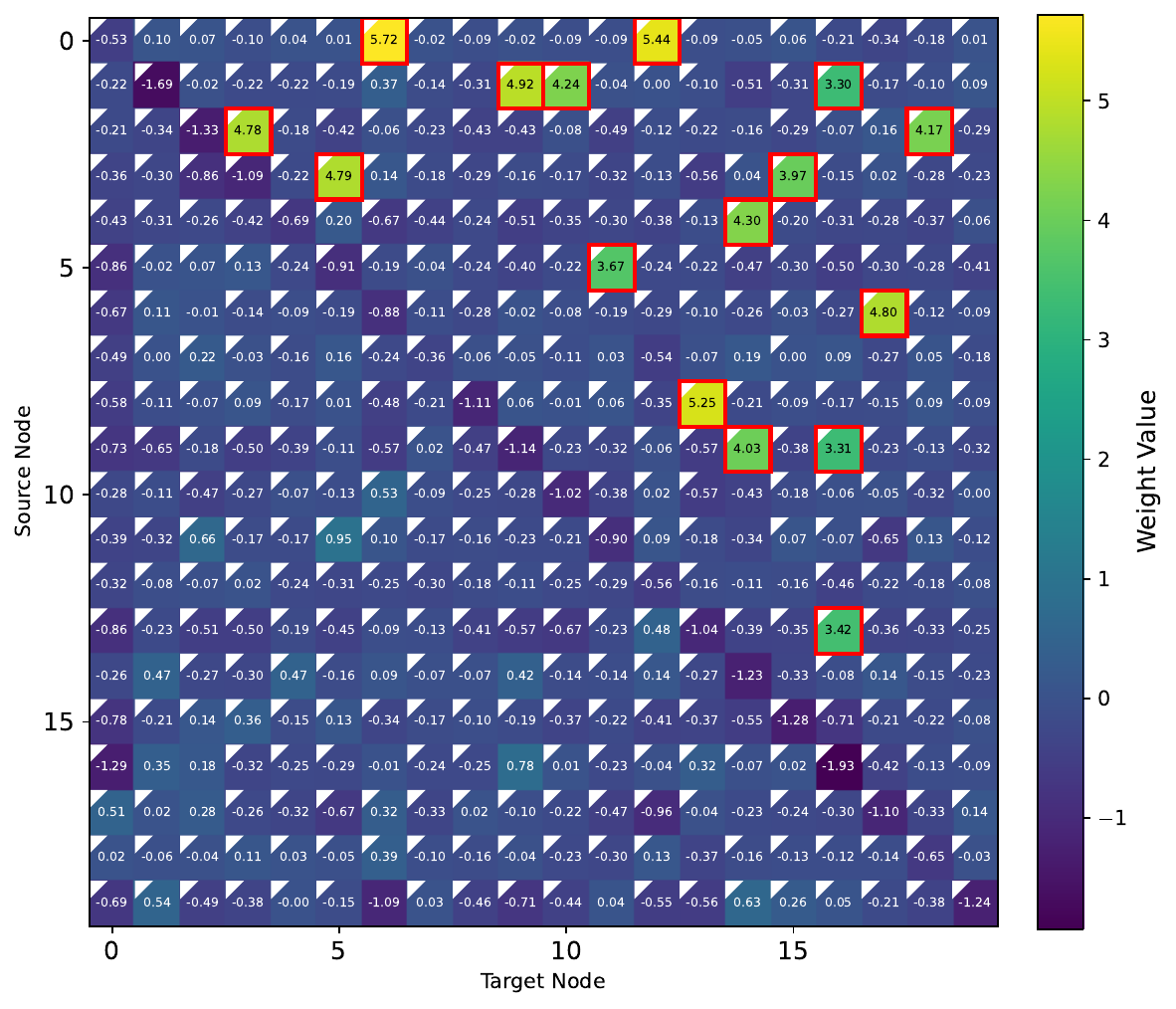} 
        \caption{Learned $\bm{W}^M$ with true adjacency}
        \label{fig:wm-heatmap-20-loss2}
    \end{subfigure}
    \hfill
    \begin{subfigure}[t]{0.48\linewidth}
        \centering
        \includegraphics[height=5cm]{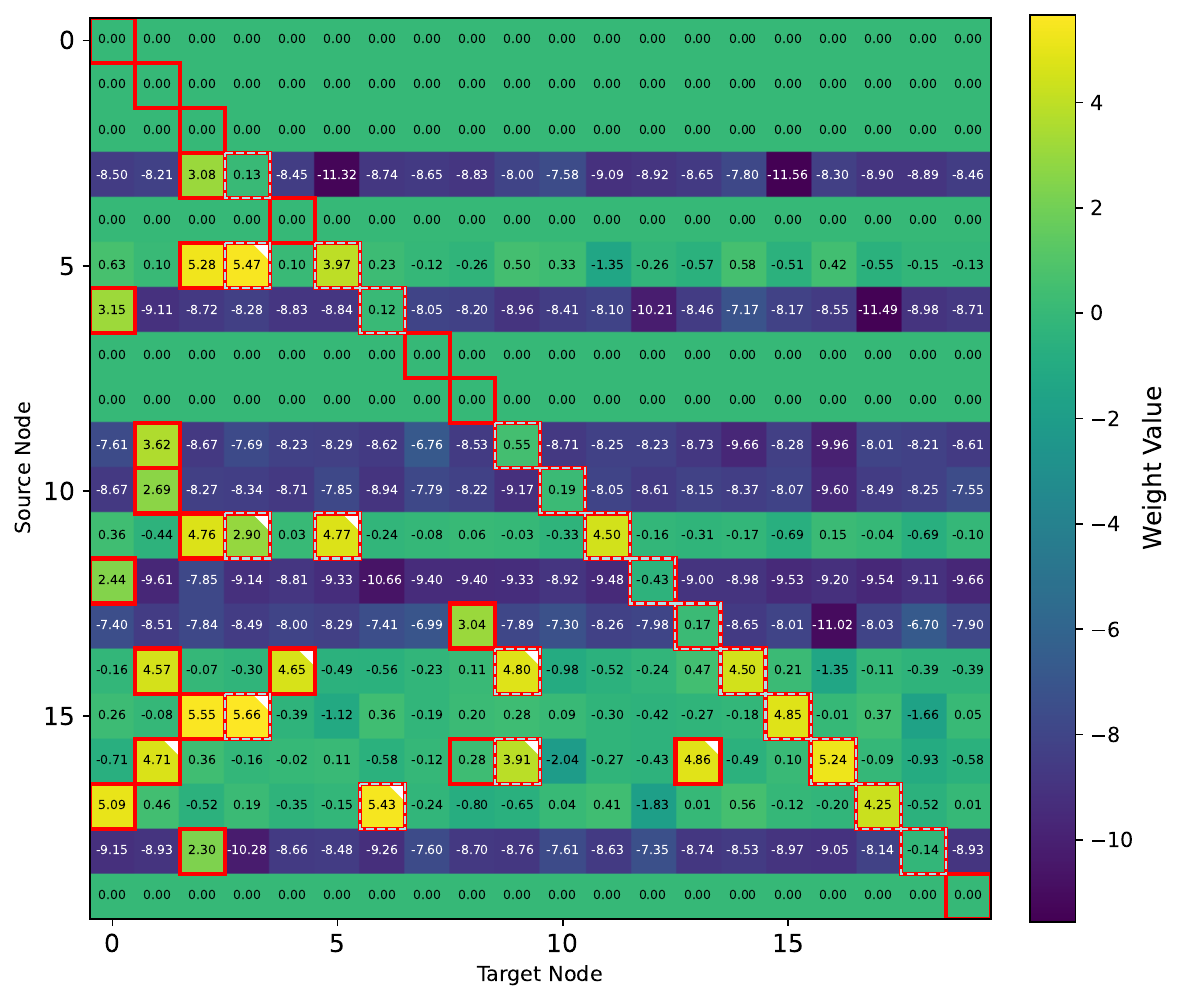} 
        \caption{Learned $\bm{W}^V$ with true reachability}
        \label{fig:wv-heatmap-20-true}
    \end{subfigure}
    
    \caption{\textbf{Weight visualizations on a 100-node graph under general 2-Token Prediction for the first 20 nodes.} 
    (a) $\bm{W}^T$ with true adjacency highlighted in red boxes. 
    (b) $\bm{W}^M$ with true adjacency in red boxes and theoretically learnable adjacency (under $\ell^{(2)}(\mathcal{D})$) marked by white triangles: extra entries are incorrect.
    (c) $\bm{W}^V$ with true reachability in red boxes, observed reachability in light dashed boxes, and theoretically learnable reachability in white: extra entries are correct and unobserved.}
    \label{fig:wt-wm-wv}
\end{figure}

Figure~\ref{fig:wt-wm-wv} shows the training results of general 2-Token Prediction. The behavior of $\bm{W}^M$ is similar to the previous analysis: the true adjacency relations learned via $\ell^{(1)}(\mathcal{D})$ remain prominent. Since $\bm{W}^T$ is not a perfect ground-truth adjacency matrix, $\bm{W}^V$ not only captures the theoretically learnable reachability relations under $\ell^{(2)}(\mathcal{D})$, but also learns additional high-weight reachability relations, which also belong to the true reachability in the graph. Table~\ref{tab:wm-wv-weights-long} shows the average weight statistics.

To explain this phenomenon in a more general way, training paths often contain sequences that differ only by additional intermediate nodes. As a result, the model tends to assign relatively high logits to tokens corresponding to these intermediate nodes, which can cause the transfer layer to learn transitions that do not exist in the true adjacency. This represents common noise in the transfer layer. At the same time, due to the structure of the paths, the transfer layer can propagate reachability through multiple steps, enabling it to capture additional correct reachability relations. Consequently, the resulting reachability relations learned by $\bm{W}^V$ include both theoretically learnable entries and other valid relations present in the graph.

This process allows the model to capture a broader set of correct reachability relations. As a result, the 2-Token Prediction models continue to achieve strong performance on the degree-3 paths in the test set, as reported in Table~\ref{tab:dag-eval-merged}.

\begin{table}[h]
\centering
\caption{ 
Average weights of different entry types in $\bm{W}^M$ and $\bm{W}^V$ in the 100-node graph under general 2-Token Prediction in the simplified model.}
\label{tab:wm-wv-weights-long}
\resizebox{0.75\linewidth}{!}{  
\begin{tabular}{clc}
\toprule
\textsc{Matrix} & \textsc{Type} & \textsc{Value} \\
\midrule
 & True adjacency & 2.57 \\
$\bm{W}^M$ & Theoretically learnable & 0.13 \\
 & Other entries & -0.11 \\
\midrule
 & Observed reachability & 0.89 \\
 & True but not observed reachability & 0.31 \\
$\bm{W}^V$ & Theoretically learnable & 0.52 \\
 & True but not observed \& theoretically learnable reachability & 0.61 \\
 & Other entries (not true reachability) & -1.21 \\
\bottomrule
\end{tabular}
}
\end{table}

\section{Blocksworld Experimental Setup}
\label{app:blocksworld_details}

The Blocksworld benchmark~\citep{valmeekam2023planning} is a well-known task in classical planning. It involves a set of colored blocks, where each block can be either placed on the table or stacked on top of another block. The objective is to plan a sequence of valid actions that transform an initial configuration into a goal configuration.

We consider a version with 4 blocks. This setup results in a complete state transition graph containing 73 nodes, where each node represents a unique, valid block configuration, and each edge \((u, v)\) denotes a legal atomic action that transitions the system from state \(u\) to state \(v\). This task can therefore be seen as an instance of the path-planning problem described in the main text, but on a fixed, predefined graph. For example, node 49 represents the state where block A is on the table, B is on C, C is on D, and D is on the table.

To simulate a more general planning scenario where the model does not see the full state space during training, we use only a small subset of state transition processes to construct the training set. Each training example corresponds to a valid path sampled from the graph. The test set is composed of randomly selected source–target state pairs, and the model is required to generate the sequence of intermediate states connecting them, i.e., a valid path.

\paragraph{Dataset Construction.}  
The training set includes all one-hop edges \((s, t) \in \mathcal{E}\) added as direct paths to ensure learning of adjacency relations. For each path length from 2 to 6, we sample \(n\) acyclic paths, where \(n\) is varied as 100, 200, 300, 400, or 500 to create training sets of different sizes. Each path is formatted as ``s t s a ... t \textbackslash n''.

The test set is fixed and consists of 5,000 randomly sampled paths with lengths greater than 1. For each training size, all models are trained on the same set of paths and evaluated on this same test set.

\section{Use of Large Language Models}
\label{app:llms}

We acknowledge the use of Large Language Models (LLMs) in the preparation of this manuscript. Their role was limited to aiding in and polishing the writing.

In adherence to the principles of transparency and academic integrity, we herein detail the auxiliary role that Large Language Models (LLMs) played in the preparation of this manuscript. We followed a strict workflow to ensure that all intellectual contributions are the original work of the human authors.
The entire substantive content of this paper—including the formulation of the research problem, the design of the theoretical framework, the execution of experiments, the collection and analysis of data, and the derivation of our conclusions—was conceived and executed exclusively by the human authors. Based on this work, we authored a complete and comprehensive first draft that fully encapsulated our research, argumentation, and findings.

Only upon the completion of this draft did we employ an LLM as a tool for linguistic refinement. Its application was strictly confined to enhancing the quality of the prose, not to generating content. Specifically, the model was utilized to check for and correct potential grammatical errors, optimize sentence structures for improved fluency and readability, and suggest more precise or varied academic terminology to enhance the overall clarity and professionalism of the text. It also assisted in ensuring the consistent use of key terms throughout the manuscript.
Critically, this refinement process remained under our rigorous supervision. Every suggestion generated by the LLM was treated as a candidate for review, not an automatic edit. Our team conducted a scrupulous evaluation and critical analysis of each proposed change, carefully judging whether it improved the language without altering our original academic intent, the nuances of our arguments, or the precision of our scientific claims. Any suggestion that could potentially introduce ambiguity or weaken our line of reasoning was not adopted.

Therefore, while we leveraged an LLM to polish the language of this manuscript, the intellectual ownership, the academic core, and the final phrasing of every sentence are the result of our independent work. The full and final responsibility for the scientific accuracy, the validity of the arguments, and the originality of this paper rests entirely with us, the human authors.

\end{document}